\documentclass{article}

\usepackage{microtype}
\usepackage{graphicx}
\usepackage{subfigure}
\usepackage{booktabs} % for professional tables

\usepackage{hyperref}

\usepackage[accepted]{icml2024}

\usepackage{amsmath}
\usepackage{amssymb}
\usepackage{mathtools}
\usepackage{amsthm}

\usepackage[capitalize,noabbrev]{cleveref}

\theoremstyle{plain}
\newtheorem{theorem}{Theorem}[section]
\newtheorem{proposition}[theorem]{Proposition}

\theoremstyle{definition}

\theoremstyle{remark}

\usepackage[textsize=tiny]{todonotes}

\usepackage{multirow}
\usepackage{multicol}

\definecolor{darkred}{RGB}{162, 0, 0}
\newcommand{\darkred}[1]{\textbf{\textcolor{darkred}{#1}}}
\definecolor{darkblue}{RGB}{4, 6, 173}
\newcommand{\darkblue}[1]{\textbf{\textcolor{darkblue}{#1}}}
\usepackage{enumitem}

\icmltitlerunning{Bounded and Uniform Energy-based Out-of-distribution Detection for Graphs}

\begin{document}

\twocolumn[
\icmltitle{ Bounded and Uniform Energy-based Out-of-distribution Detection for Graphs
}

% It is OKAY to include author information, even for blind
% submissions: the style file will automatically remove it for you
% unless you've provided the [accepted] option to the icml2024
% package.

% List of affiliations: The first argument should be a (short)
% identifier you will use later to specify author affiliations
% Academic affiliations should list Department, University, City, Region, Country
% Industry affiliations should list Company, City, Region, Country

% You can specify symbols, otherwise they are numbered in order.
% Ideally, you should not use this facility. Affiliations will be numbered
% in order of appearance and this is the preferred way.
\icmlsetsymbol{equal}{*}

\begin{icmlauthorlist}
\icmlauthor{Shenzhi Yang}{suda}
\icmlauthor{Bin Liang}{cuhk}
\icmlauthor{An Liu}{suda}
\icmlauthor{Lin Gui}{kcl}
\icmlauthor{Xingkai Yao}{suda}
\icmlauthor{Xiaofang Zhang}{suda}
% \icmlauthor{Firstname7 Lastname7}{comp}
% %\icmlauthor{}{sch}
% \icmlauthor{Firstname8 Lastname8}{sch}
% \icmlauthor{Firstname8 Lastname8}{yyy,comp}
%\icmlauthor{}{sch}
%\icmlauthor{}{sch}
\end{icmlauthorlist}

\icmlaffiliation{suda}{School of Computer Science and Technology, Soochow University, Suzhou, China}
\icmlaffiliation{cuhk}{Department of Systems Engineering and Engineering Management, The Chinese University of Hong Kong, Hong Kong, China}
\icmlaffiliation{kcl}{Department of Informatics, King's College London, London, UK}

\icmlcorrespondingauthor{Bin Liang}{bliang@se.cuhk.edu.hk}
\icmlcorrespondingauthor{Xiaofang Zhang}{xfzhang@suda.edu.cn}

% You may provide any keywords that you
% find helpful for describing your paper; these are used to populate
% the "keywords" metadata in the PDF but will not be shown in the document
\icmlkeywords{Machine Learning, ICML}

\vskip 0.3in
]

% this must go after the closing bracket ] following \twocolumn[ ...

% This command actually creates the footnote in the first column
% listing the affiliations and the copyright notice.
% The command takes one argument, which is text to display at the start of the footnote.
% The \icmlEqualContribution command is standard text for equal contribution.
% Remove it (just {}) if you do not need this facility.

%\printAffiliationsAndNotice{}  % leave blank if no need to mention equal contribution
% \printAffiliationsAndNotice{\icmlEqualContribution} % otherwise use the standard text.
\printAffiliationsAndNotice{} % otherwise use the standard text.

\begin{abstract}\label{abstract}
Given the critical role of graphs in real-world applications and their high-security requirements, improving the ability of graph neural networks (GNNs) to detect out-of-distribution (OOD) data is an urgent research problem. The recent work GNNSAFE \citep{wu2023energy} proposes a framework based on the aggregation of negative energy scores that significantly improves the performance of GNNs to detect node-level OOD data. However, our study finds that score aggregation among nodes is susceptible to extreme values due to the unboundedness of the negative energy scores and logit shifts, which severely limits the accuracy of GNNs in detecting node-level OOD data. In this paper, we propose NODESAFE: reducing the generation of extreme scores of nodes by adding two optimization terms that make the negative energy scores bounded and mitigate the logit shift. Experimental results show that our approach dramatically improves the ability of GNNs to detect OOD data at the node level, e.g., in detecting OOD data induced by Structure Manipulation, the metric of FPR95 (lower is better) in scenarios without (with) OOD data exposure are reduced from the current SOTA by \textbf{28.4}\% (\textbf{22.7}\%). The code is available via \href{https://github.com/ShenzhiYang2000/NODESAFE-Bounded-and-Uniform-Energy-based-Out-of-distribution-Detection-for-Graphs}{https://github.com/ShenzhiYang2000/NODESAFE}.
\end{abstract}

\section{Introduction}\label{sec-Introduction}

%%%%%%%%%%%%%%%%%%%%%%%%%%%%%%%%%%%%%%%%%%%%%%%
\begin{figure}[!t]
	\centering
   \subfigure[Out-of-Distribution]{
    % \begin{minipage}[b]{0.47\linewidth}
		\includegraphics[width=0.45\linewidth]{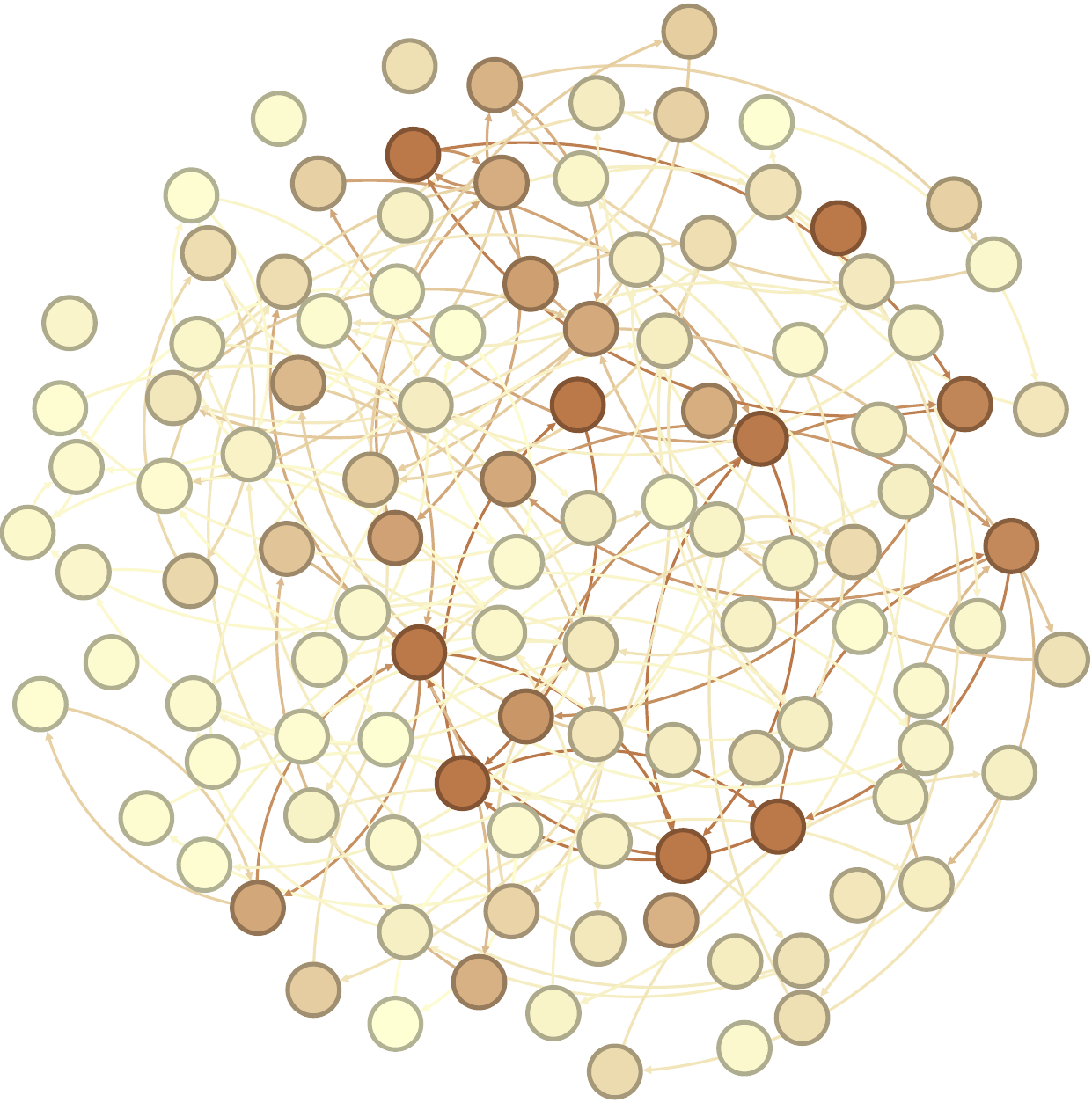}
	% \end{minipage}
 }
  \subfigure[In-Distribution]{
	% \begin{minipage}[b]{0.47\linewidth}
		\includegraphics[width=0.45\linewidth]{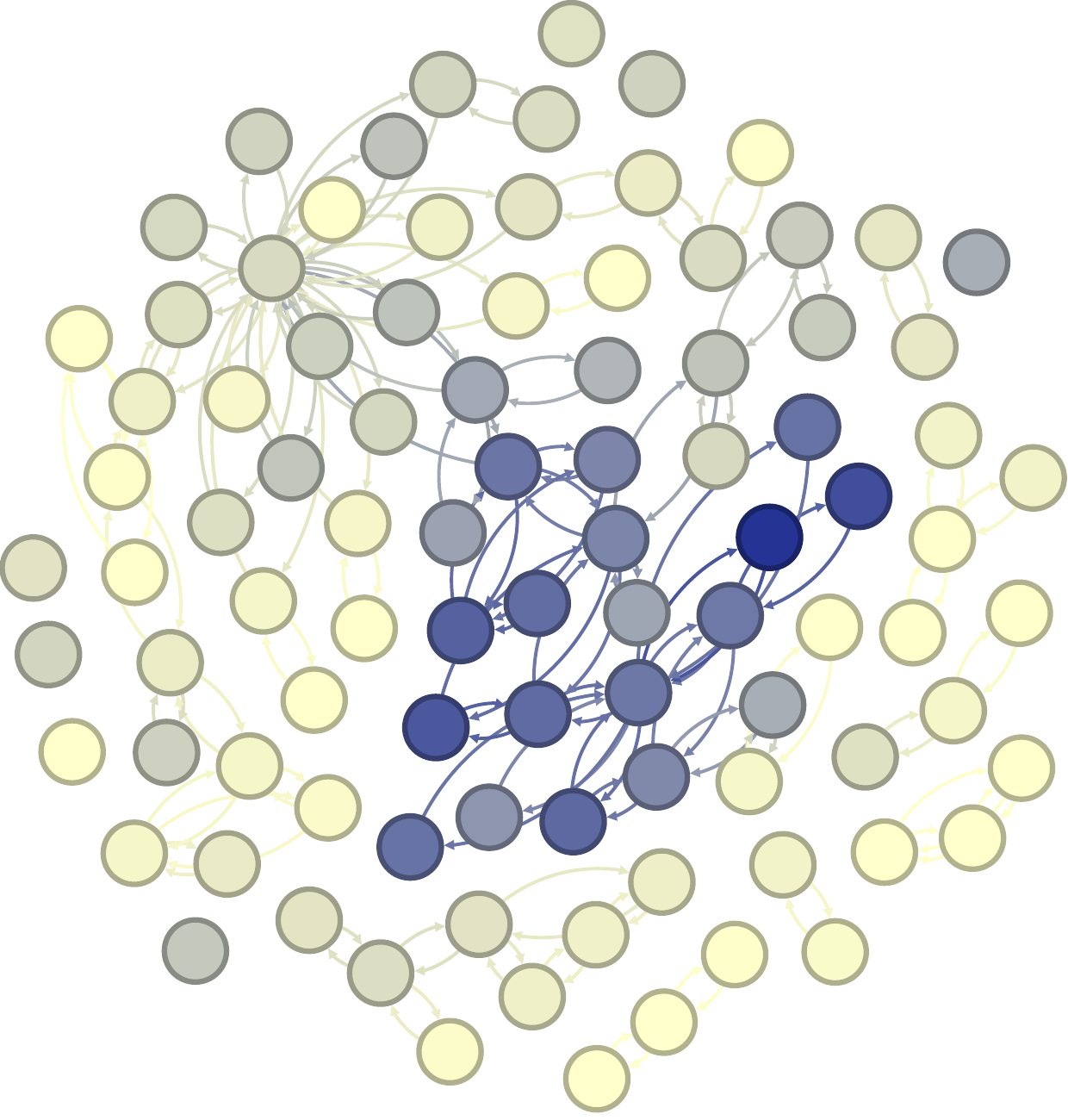}
	% \end{minipage}
 }

   \subfigure[Score Variance Reduction]{
    % \begin{minipage}[b]{0.47\linewidth}
		\includegraphics[width=0.7\linewidth]{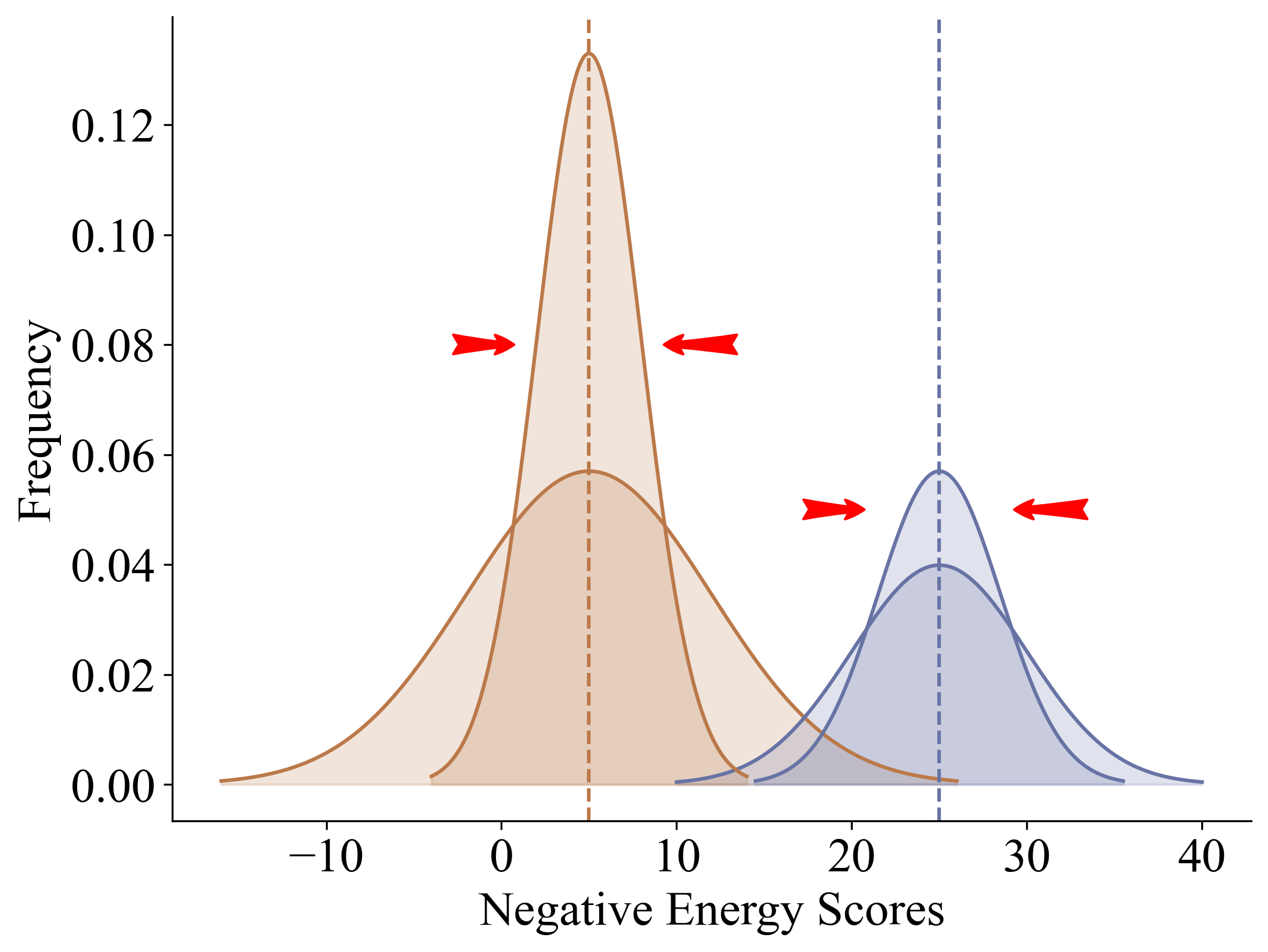}
	% \end{minipage}
 }
    \caption{Visualization of negative energy scores of different nodes. (a) OOD data generated by Structure Manipulation \citep{wu2023energy}, dark colors indicate higher scores, and arrows between nodes indicate the direction of score aggregation. (b) ID dataset for Cora, dark colors indicate lower scores. (c) Score frequency density plots show our motivation: lowering the variance of scores can better categorize ID and OOD data.}
    \label{F-Motivation}
\end{figure}

The graph is an essential data structure widely used in various real-world domains, such as knowledge graphs~\cite{baek2020learning}, social networks~\cite{fan2019graph}, point clouds~\cite{shi2020point}, and chemical analysis~\cite{de2018molgan}. This has attracted a great deal of interest from researchers, who have developed several advanced models, such as the GCN\citep{kipf2016semi}, GAT\citep{velivckovic2017graph}, JKNet \citep{xu2018representation}, MixHop \citep{abu2019mixhop}, etc. However, there are concerns regarding the security issues that may arise when applying graph neural networks. During training, it is assumed that both the training and testing data are in-distribution (ID). However, this assumption may not always hold, as data can change at any time. When graph neural networks encounter out-of-distribution (OOD) data, they may make incorrect predictions, potentially leading to serious consequences. Therefore, it is crucial to explore methods that enable graph neural networks to detect OOD data effectively. Various methods \citep{hendrycks2016baseline,liang2017enhancing,lee2018simple,hendrycks2018deep,liu2020energy}  have been developed to identify out-of-distribution (OOD) data, with a primary focus on i.i.d. samples, such as images. Some work \citep{li2022graphde,bazhenov2022towards,liu2023good,guo2023data,ding2023sgood} has also been done on OOD detection at the graph level in the graph domain, but these methods cannot be directly applied to OOD detection at the node level since each graph can still be viewed as i.i.d samples while nodes are not. In addition, there is a relative lack of methods \citep{zhao2020uncertainty,stadler2021graph} specifically designed to detect node-level OOD data on graphs.  
%\citep{zhao2020uncertainty} designed a graph-based kernel Dirichlet distribution estimation (GKDE) method. \citep{stadler2021graph}  proposed graph posterior networks (GPNs) that explicitly perform Bayesian a posteriori updating for predicting interdependent nodes. 
Recently, inspired by the label propagation algorithm \citep{raghavan2007near},  \citep{wu2023energy} propose a node-level OOD detection method called GNNSAFE, which aggregates negative energy scores and significantly improves the accuracy of OOD node detection. It is considered the most efficient method for detecting node-level OOD data on graphs. 

However, we have observed that the aggregation of negative energy scores among nodes suffers from excessive variance in both ID and OOD data. This is because the node's negative energy score updates are easily affected by the extreme scores of neighboring nodes. For instance, as illustrated in Figure \ref{F-Motivation} (a) and (b), if a node has a high (resp. low) score, score propagation will tend to increase (resp. decrease) the scores of its neighboring nodes. To improve OOD detection at the node level, our work explores how to reduce the generation of extreme scores, as this will lessen the accuracy of OOD detection.

We find two main reasons exist for generating extreme node scores. Firstly, the negative energy scores are unbounded, which is the root cause of generating extreme scores. Secondly, the logit shift may arise during the node classification optimization, which is the direct cause. Logit shift means that adding an arbitrary constant to the logit will not affect the model's supervised classification performance but will cause the negative energy scores to change. 

Our method, NODESAFE, addresses the issues by incorporating two combinable optimization objectives. To address the issue of unbounded negative energy scores, we add an optimization term to minimize the variance of the logit 2-norms of different samples to make the negative energy scores bounded. To tackle the issue of logit shift, we aim to reduce the objective of the summation variance of logit. This approach helps converge the negative energy scores of both ID and OOD data to their respective means, as shown in Fig. \ref{F-Motivation}(c). This, in turn, facilitates better separation of ID and OOD data. In summary, our contributions are as follows:

\begin{enumerate}[nosep, topsep=0pt, leftmargin=*]
    \item We reveal how extreme scores during score aggregation affect the node-level OOD detection performance of GNNs and propose NODESAFE with two constraints to curb the generation of extreme scores. To our knowledge, we are the first to enhance GNN's OOD detection capability from both the upper and lower bounds of the 2-norm of logits.
    \item  We have conducted extensive experiments on a large number of real-world datasets and multiple OOD data generation scenarios, which demonstrates that NODESAFE significantly improves the node-level OOD detection performance of GNNs in various scenarios, e.g., in detecting OOD data generated by Structure Manipulation \citep{wu2023energy}, we have reduced the FPR95 (lower is better) metrics of the scenarios without (with) OOD data exposure by \textbf{28.4}\% (\textbf{22.7}\%) compared to the current SOTA method GNNSAFE. 
    \item  We have performed a sound theoretical analysis of NODESAFE, which is simple but clearly explains the reason for its remarkable effect.
  \end{enumerate}

\section{Background}\label{sec-Background}
This section introduces the background knowledge necessary for our work. Section \ref{subsec-Graph_Neural_Network} presents the graph neural network (GNN) definition. Section \ref{subsec-OOD_Detection} describes the Out-of-Distribution (OOD) problem. Section \ref{subsec_Energy-based Model} introduces energy-based OOD methods, including the current state-of-the-art node-level OOD detection method GNNSAFE \citep{wu2023energy}.

\subsection{Graph Neural Network}\label{subsec-Graph_Neural_Network}
Let $\mathcal{G}=\{\mathcal{V}, \mathcal{E}\}$ denote an undirected graph, where $\mathcal{V}$ and $\mathcal{E}$ are the sets of nodes and edges, respectively, ${\boldsymbol X}_v \in \mathbb {R}^{|\mathcal{V}|\times d_v}$ denote the matrix of features of the nodes. Here, the representation of a node $v$ can be defined as $\mathbf{h}_v$. Graph neural networks (GNNs) aim to update the representation of the given graph $\mathcal{G}$ by leveraging its topological structure. For the representation $\mathbf{h}_v$ of node $v$, its propagation of the $k$-th layer GNN is represented as:
\begin{equation}\label{equa-gnns}
\begin{array}{cl}
   \mathbf{h}_v^{(k)}&=f(\mathbf{h}_v^{(k-1)};\theta)    \\
            &=\mathrm{UP}^{(k)}\big(\mathrm{AGG}^{(k)}\big( \mathbf{h}_u^{(k-1)}:\forall u \in \mathcal{N}(v)\cup v \big) \big) 
\end{array} 
\end{equation}
where $f(\cdot;\theta)$ denotes the GNN encoder, $\theta$ represents all trainable parameters of GNN encoder. $\mathrm{AGG}(\cdot)$ denotes a trainable function that aggregates messages from the neighbors of node $v$, $\mathcal{N}(v)$ represents the set of neighbors. $\mathrm{UP}(\cdot)$ denotes a trainable function that updates the representation of node $v$ with the current representation of $v$ and the aggregated vector. In the node classification task, the $\mathbf{z}_v$ obtained after $\mathbf{h}_v$ passes through the classification layer is usually called \textit{logit}. After logit passes through softmax, the probabilities $p(y|\mathbf{z})$ of different categories are obtained as follows:
$
    p(y|\mathbf{z}) = \frac{e^{z_y}}{\sum_{c=1}^C e^{z_c}}
$ 
where $C$ denotes the number of classes.

\subsection{Problem Statement: Out-of-distribution Detection}\label{subsec-OOD_Detection}
Within the realm of machine learning, the concept of out-of-distribution (OOD) detection pertains to the identification of instances $\mathbf{x} \in \mathcal{X}$ that lie outside the data distribution $\mathcal{D}_{in}$ (ID) upon which the model is trained. OOD detection entails the utilization of scoring functions to ascertain confidence levels for each instance, as outlined below:
\begin{equation}\label{equa-ood}
    g(\mathbf{x})=\left\{
\begin{array}{ll}
\mathrm{in}
&   S(\mathbf{x}) \geq \gamma \\
\mathrm{out}
&  S(\mathbf{x}) \leq \gamma \\
\end{array} \right.
\end{equation}
where $S(\mathbf{x})$ represents a scoring function, with $\gamma$ typically chosen to guarantee a high percentage, such as 95\%, of ID data that is correctly classified. In line with convention, instances with elevated scores derived from the function $S(\mathbf{x})$ are classified as ID, while those with lower scores are classified as OOD.

\subsection{Energy-based OOD Detection Method} \label{subsec_Energy-based Model}
\paragraph{OOD detection in scenarios where samples are independent.} An energy-based model (EBM)\citep{lecun2006tutorial,liu2020energy,wu2023energy} establishes a connection between classification models and energy models by constructing a function $E(\mathbf{x}): \mathbb{R}^D \rightarrow \mathbb{R}$. This function maps each point $\mathbf{x} \in \mathbb{R}^D$ (such as logit $\mathbf{z}$) to a nonprobabilistic scalar known as energy. The mapping is represented as follows:
\begin{equation}\label{equa-Energy}
    E(\mathbf{z}) = - \mathrm{log}\sum_{c=1}^C e^{\mathrm{z}_c}
\end{equation}
 In the context of classification scenarios, reducing the negative log-likelihood function serves to decrease the energy of data within the distribution, as illustrated below:
\begin{equation}\label{equa-L_nll}
\begin{aligned}
    \mathcal{L}_{nll} &= \mathbb{E}_{(\mathbf{z}, y) \sim \mathcal{D}_{in}}\big( -\mathrm{log} p(y | \mathbf{z}) \big)\\
    &= \mathbb{E}_{(\mathbf{z}, y) \sim \mathcal{D}_{in}} \big( -\mathrm{z}_y + \mathrm{log} \sum_{c=1}^C e^{\mathrm{z}_c}  \big)
\end{aligned}
\end{equation}
 By defining the energy $E(\mathbf{z},y) = -\mathrm{z}_y$, we can rewrite the NLL loss as:
\begin{equation}\label{equa-L_nll_rewrite}
\begin{array}{cl}
    \mathcal{L}_{nll} = \mathbb{E}_{(\mathbf{z}, y) \sim \mathcal{D}_{in}} \big( E(\mathbf{z},y) + \mathrm{log} \mathop{\sum}\limits_{c \neq y}  e^{-E(\mathbf{z}, c)}  \big)\\
\end{array}
\end{equation}
Eq. \ref{equa-L_nll_rewrite} shows that $E(\mathbf{z})$ decreases with $\mathcal{L}_{nll}$ during the optimization process. However, to comply with convention, i.e., Eq. \ref{equa-ood}, we use negative energy $-E(\mathbf{z})$ as the scoring function.

In the case of OOD data exposure, past work\citep{liu2020energy,wu2023energy} expands the energy interval between ID and OOD data with two squared hinge losses as follows:
\begin{equation}\label{equa-L_Energy}
\begin{aligned}
    \mathcal{L}_{reg} &= \mathbb{E}_{(\mathbf{z}_{in}, y) \sim \mathcal{D}_{in}}\big( max(0, E(\mathbf{z}_{\mathrm{in}})-m_{\mathrm{in}})^2 \big)\\
    &+ \mathbb{E}_{(\mathbf{z}_{out}, y) \sim \mathcal{D}_{out}}\big( max(0, m_{\mathrm{out}}-E(\mathbf{z}_{\mathrm{out}}))^2 \big)
\end{aligned}
\end{equation}
where $m_{\mathrm{in}}$ and $m_{\mathrm{out}}$ denote the separate margin hyperparameters respectively. The overall optimization objectives for fine-tuning the model using exposed OOD data with a trade-off hyper-parameter $\alpha$ are as follows:
% \begin{equation}
%     \mathop{\mathrm{min}}\limits_\theta \quad \mathbb{E}_{(\mathbf{z}, y) \sim \mathcal{D}_{in}}\big( -\mathrm{log} p(y | \mathbf{z}) \big) + \alpha \cdot \mathcal{L}_{Energy}
% \end{equation}
\begin{equation} \label{equa-L_OOD }
\mathcal{L}_{OOD} =  \mathcal{L}_{nll} + \alpha \cdot \mathcal{L}_{reg}
\end{equation}

% \subsection{Energy-based OOD Detection: GNNSAFE} \label{subsec-Energy_Aggregation}
\paragraph{OOD detection in scenarios where samples are not independent.} \label{subsec-Energy_Aggregation}

Based on the energy-based approach \citep{liu2020energy} for OOD detection on independent samples and inspired by label propagation, GNNSAFE \citep{wu2023energy} introduces negative energy score aggregation to iteratively propagate the scores among interconnected nodes to amplify the difference in scores between ID data and OOD data:
\begin{equation} \label{equa-Energy_Agg}
    \mathbf{E}^{(k)} = \eta\mathbf{E}^{(k-1)} + (1-\eta)\mathcal{D}^{-1}\mathcal{A}\mathbf{E}^{(k-1)}
\end{equation}
where $\mathcal{D}$ and $\mathcal{A}$ respectively represent the degree matrix and adjacency matrix. $\eta \in [0,1]$ is a hyper-parameter that controls the intensity of energy fusion. $k$ is the number of iteration hops for energy aggregation. 

However, GNNSAFE does not account for the impact of extreme values during energy score aggregation. For OOD nodes, extremely high energy scores are outliers, while for ID nodes, extremely low energy scores are outliers. We provide the following proof:
For OOD node $i$, if $\frac{\sum_j \eta_{ij}{E}^{(k-1)}_j}{\sum_j \eta_{ij}} > {E}^{(k-1)}_i$ as the neighboring node $j$ with extremely high energy score existing, we can re-write Eq.  \ref{equa-Energy_Agg} as the scalar-form for each instance:
\begin{equation}
\begin{aligned}
    {E}^{(k)}_i &= \eta {E}^{(k-1)}_i + (1-\eta) \frac{\sum_j \eta_{ij}{E}^{(k-1)}_j}{\sum_j \eta_{ij}} \\
    &> \eta {E}^{(k-1)}_i + (1-\eta) {E}^{(k-1)}_i\\
    &> {E}^{(k-1)}_i
\end{aligned}
\end{equation}
The derivation above demonstrates that OOD nodes with extremely high energy scores will increase the energy scores of neighboring nodes, thereby affecting OOD detection. Conversely, ID nodes with extremely low energy scores will decrease the energy scores of neighboring nodes, similarly affecting OOD detection. Therefore, we need to restrict the occurrence of extreme energy scores in both OOD and ID. 

\section{Method}\label{sec-Method}
This section describes our improvements to the aggregation of negative energy scores. Section \ref{subsec-Motivation} describes our motivation, and Section \ref{subsec-bou_and_uni_agg} presents our proposed method, NODESAFE.

\subsection{Motivation} \label{subsec-Motivation}
In the following discussion, we will delve into the reasons for the large variance in the respective scores of the ID data and the OOD data.
% \begin{proposition} \label{prop-logitshift}
%     \textbf{Logit Shift}: Consider the softmax function denoted by $\sigma$ for the softmax cross-entropy loss. Given any constant $s$, if $\mathrm{c=argmax_i(z_i)}$, then the identity $\sigma_{\mathrm{c}}(\mathbf{z} + s) = \sigma_{\mathrm{c}}(\mathbf{z})$ always holds. Furthermore, when $s \leq 0$, we have $-\infty \leq -E(\mathbf{z}+s) \leq -E(\mathbf{z})$, and when $s \geq 0$, $+\infty \geq -E(\mathbf{z}+s) \geq -E(\mathbf{z})$.
% \end{proposition}

\begin{proposition} \label{prop-logitshift}
    \textbf{Logit Shift}: Consider the softmax function denoted by $\sigma$ for the softmax cross-entropy loss. Given any constant $s$, if $\mathrm{c=argmax_i(z_i)}$, then the identity $\sigma_{\mathrm{c}}(\mathbf{z} + s) = \sigma_{\mathrm{c}}(\mathbf{z})$ always holds.
\end{proposition}

\begin{proposition} \label{prop-unbounded}
    \textbf{Boundlessness of energy}: Given any constant $s$, if $s \leq 0$, we have $-\infty < -E(\mathbf{z}+s) \leq -E(\mathbf{z})$, and if $s > 0$, $+\infty > -E(\mathbf{z}+s) > -E(\mathbf{z})$.
\end{proposition}
We give the proofs of Proposition \ref{prop-logitshift} and \ref{prop-unbounded} in the appendix \ref{Appendix-Proof-Logitshift} and \ref{Appendix-Proof-unbounded}, respectively. From Proposition \ref{prop-logitshift}, we observe that there may be logit shifts even if samples can be correctly classified. While logit shifts do not affect sample classification, they significantly affect negative energy scores. Due to the unbounded nature of the negative energy scores, as demonstrated in Proposition \ref{prop-unbounded}, when $s$ is extremely small or large, they exhibit minimal or high scores, respectively, for samples from ID or OOD data. During score aggregation on the graph, scores that deviate significantly from the mean can influence the scores of neighboring nodes, thereby increasing the variance of scores for samples within ID and OOD, respectively.
Consequently, this variability impacts the performance of OOD detection by GNNs. Therefore, we posit that reducing the variance of sample scores can enhance OOD detection performance on the graph, as illustrated in Figure \ref{F-Motivation}(c). Our approach aims to mitigate extreme score values, fostering a bounded and more uniform distribution of scores for data within ID and OOD, respectively.

\subsection{NODESAFE: Bounded and Uniform Negative Energy Score Aggregation} \label{subsec-bou_and_uni_agg}
As analyzed above, the root reason for generating extreme scores is that they lack upper and lower bounds. Therefore, we address this issue by constraining the 2-norm of the logit to create upper and lower bounds for negative energy scores, as follows:
\begin{gather}\label{equa-L_bound}
    M_{\mathrm{norm}} = \frac{1}{|\mathcal{V}|} \sum_{v \in \mathcal{V}} \lVert \mathbf{z}_v \rVert_2 \\
    \mathcal{L}_{\mathrm{bound}} = M_{\mathrm{norm}}^{-1}\cdot \mathbb{E}_{\mathbf{z}\sim \mathcal{D}_{tr}} \big( ( \lVert \mathbf{z} \rVert_2 - \frac{1}{|\mathcal{V}|} \sum_{v \in \mathcal{V}} \lVert \mathbf{z}_v \rVert_2)^2 \big)
\end{gather}
where $\lVert \mathbf{z}_v \rVert_2 = \sqrt{\sum_{i \in v} \mathrm{z}_i}$ denotes the 2-norm of the logit $\mathbf{z}_v$. To ensure numerical stability during the optimization process, we use the ratio of variance to mean as the optimization objective; the smaller, the better.
\begin{proposition}\label{prop-Energy_bound}
    For logit vectors with equal 2-norms of $M_{\mathrm{norm}}$, their upper bound for negative energy scores $(-\mathbf{E})^{upper}=\mathrm{log}C+\frac{M_{\mathrm{norm}}}{\sqrt{C}}$ , and the lower bound  $(-\mathbf{E})^{lower}=\mathrm{log}C-\frac{M_{\mathrm{norm}}}{\sqrt{C}}$.
\end{proposition}
We give the proof of the Proposition \ref{prop-Energy_bound} in the appendix \ref{Appendix-proof-Energy_bound}. Proposition \ref{prop-Energy_bound} shows that when the logit of all samples is distributed on a hypersphere with a radius of $M_{\mathrm{norm}}$, it can effectively prevent the generation of high and minimal negative energy scores. For the logit shift, we take the variance of the logit summation as the optimization objective; the smaller, the better, as follows:
\begin{gather}\label{equa-L_uniform}
    M_{\mathrm{sum}} = \frac{1}{|\mathcal{V}|} \sum_{v \in \mathcal{V}} \Sigma(\mathbf{z}_v)  \\
    \mathcal{L}_{\mathrm{uniform}}^v = ( \Sigma(\mathbf{z}_v) - \frac{1}{|\mathcal{V}|} \sum_{i \in \mathcal{V}} \Sigma(\mathbf{z}_i))^2\\
    \mathcal{L}_{\mathrm{uniform}} =  \sum_{d \in (\mathrm{in}, \mathrm{out})} M_{\mathrm{sum}}^{-1} \cdot \mathbb{E}_{\mathbf{z}_v \sim \mathcal{D}_{d}} (\mathcal{L}_{\mathrm{uniform}}^v)
\end{gather}
where $\Sigma(\mathbf{z}_v)$ denotes $ \sum_{i} \mathrm{z}_v^i$. Note that $M_{\mathrm{norm}}$ and $M_{\mathrm{sum}}$ serve solely as scaling coefficients and do not participate in gradient propagation. Unlike $\mathcal{L}_{\mathrm{bound}}$, $\mathcal{L}_{\mathrm{uniform}}$ is calculated separately for ID and OOD samples (if there is OOD exposure). In Appendix \ref{Appendix-sec-more_analys_on_Lub}, we provide a theoretical analysis of $\mathcal{L}_{\mathrm{bound}}$ and $\mathcal{L}_{\mathrm{uniform}}$, computing their gradients and demonstrating their impact on the logit $\mathbf{z}$, as well as their enhancement of the OOD detection performance of the model. The final optimization objective for our method is composed as follows:
\begin{gather}\label{equa-All}
    \mathcal{L}_{UB} = (\lambda_1 \cdot \mathcal{L}_{\mathrm{uniform}} + (1-\lambda_1) \cdot \mathcal{L}_{\mathrm{bound}}) \\
    \mathcal{L}_{ALL} = \mathcal{L}_{OOD} + \lambda_2 \cdot \mathcal{L}_{UB}
\end{gather}
where $\lambda_1$ and $\lambda_2$ are hyper-parameters.

\begin{table*}[!t]
% \scriptsize
\centering
\caption{Out-of-distribution detection performance measured by AUROC ($\uparrow$) / AUPR ($\uparrow$) / FPR95 ($\downarrow$) on datasets \textit{Cora},
\textit{Citeseer}, \textit{Pubmed} with three OOD types (Structure Manipulation, Feature Interpolation, Label
Leave-out ) \citep{wu2023energy}. We use the same GCN as GNNSAFE\citep{wu2023energy} as the backbone in all experiments.}\label{tabel-Cora_C_P}
\resizebox{\linewidth}{!}{
% \scalebox{0.95}{
\begin{tabular}{c|c|c|ccc|ccc|ccc}
\hline
\hline
\multirow{2}*{Metric} &\multirow{2}*{Model}  &OOD &\multicolumn{3}{c|}{Cora} &\multicolumn{3}{c|}{Citeseer} &\multicolumn{3}{c}{PubMed}\\
~ &~ &Expo &S	&F	&L	&S	&F	&L &S	&F	&L\\
\hline
\hline
\multirow{14}{*}{\rotatebox{90}{FPR95 ($\downarrow$)}}&MSP	&No	&87.30	&64.88	&34.99		&85.03	&71.27	&51.97		&84.08	&69.38	&46.19\\
~ &ODIN	&No	&100.00	&100.00	&100.00		&100.00	&100.00	&100.00		&100.00	&100.00	&100.00\\
~ &Mahalanobis	&No	&98.19	&99.93	&90.77		&99.13	&99.73	&86.32		&97.59	&84.93	&78.21\\
~ &Energy	&No	&88.74	&65.81	&41.08		&87.59	&69.67	&38.76		&78.90	&62.47	&45.14\\
~ &GKDE	&No	&84.34	&68.24	&88.95		&93.71	&71.22	&50.61		&81.52	&68.56	&69.52\\
~ &GPN	&No	&76.22	&56.17	&37.42		&78.26	&73.14	&41.37		&80.33	&61.79	&50.23\\
~ &GNNSAFE&	No	&73.15&	38.92	&\darkblue{30.83}	&	\darkblue{74.72}	&\darkblue{68.83}	&\darkblue{36.53}	&	\darkblue{44.64}	&\darkblue{33.89}	&36.49\\
~ &GNNSAFE w/ $\mathcal{L}_{LN}$ &	No	&\darkblue{61.04}	&\darkblue{38.44}&	34.99		&76.35&72.35	&37.32	&	75.10	&49.93	&\darkblue{32.29}\\
~ &NODESAFE (ours)&No	&\darkred{25.63}	&\darkred{23.08}	&\darkred{29.41}		&\darkred{57.89}	&\darkred{42.47}	&\darkred{29.30}	&	\darkred{23.80}	&\darkred{22.01}	&\darkred{25.01}\\

\cline{2-12}							

~ &OE	&Yes&	95.31	&83.79&	46.55	&	95.37	&81.09&	45.99		&83.52	&74.58&	60.30\\
~ &Energy FT	&Yes&	67.73	&47.53	&37.83	&	76.44&	\darkblue{64.08}&	31.60	&	92.04	&90.00	&\darkblue{25.59}\\
~ &GNNSAFE++	&Yes	&53.51&	\darkblue{27.73}	&34.08	&	\darkblue{70.72}	&72.98	&\darkblue{29.30}	&	\darkblue{34.43}	&\darkblue{26.30}&	33.63\\
~ &GNNSAFE++   w/  $\mathcal{L}_{LN}$	&Yes&	\darkblue{51.99}	&32.72&	\darkblue{28.40}	&	74.81&	75.47	&30.55	&	91.58	&86.17	&27.81\\
~ &NODESAFE++ (ours)	&Yes	&\darkred{23.34}&	\darkred{14.73}&	\darkred{22.52}&	\darkred{52.60}	&\darkred{40.49}&	\darkred{29.04}	&	\darkred{14.52}	&\darkred{24.45}	&\darkred{23.81}\\

\hline
\hline

\multirow{14}{*}{\rotatebox{90}{AUROC ($\uparrow$)}} &MSP	&No	&70.90	&85.39	&91.36		&66.34	&78.32	&88.42		&74.31	&83.28	&85.71\\
~ &ODIN	&No&	49.92	&49.88	&49.80	&	49.23	&49.86&	51.33	&	49.76&	49.67	&56.24\\
~ &Mahalanobis	&No	&46.68	&49.93	&67.62	&	45.26&	49.92&	53.46	&	55.28	&69.12	&75.77\\
~ &Energy	&No	&71.73	&86.15	&91.40		&65.62	&79.19	&89.98	&	74.33	&84.16	&86.81\\
~ &GKDE	&No	&68.61	&82.79	&57.23		&61.48	&74.68	&82.69		&74.02	&82.25	&83.36\\
~ &GPN	&No	&77.47	&85.88	&90.34		&70.55	&78.46	&85.65		&74.96&	82.56	&86.51\\
~ &GNNSAFE	&No	&87.52	&\darkblue{93.44}	&92.80		&79.79	&\darkblue{83.46}	&90.01		&87.52	&\darkblue{94.28}	&88.02\\
~ &GNNSAFE w/ $\mathcal{L}_{LN}$ &	No	&\darkblue{88.33}	&93.26&	\darkblue{93.50}		&\darkblue{84.67}	&88.11	&\darkblue{90.47}		&\darkblue{89.31}	&92.07&	\darkblue{91.12}\\
~ &NODESAFE (ours) &No&	\darkred{94.07}	&\darkred{95.30}	&\darkred{93.80}		&\darkred{88.40}	&\darkred{90.41}&	\darkred{91.66}		&\darkred{94.13}	&\darkred{95.97}&	\darkred{93.80}\\

\cline{2-12}									

~ &OE	&Yes	&67.98	&81.83&	89.47	&	58.74	&72.06&	89.44	&	74.41	&82.34	&81.97\\
~ &Energy FT	&Yes	&75.88	&88.15	&91.36		&68.87&	79.23	&91.34		&73.54	&78.95	&\darkblue{91.83}\\
~ &GNNSAFE++	&Yes	&\darkblue{90.62}	&\darkblue{95.56}	&92.75		&82.43	&83.27	&\darkblue{91.57}		&\darkblue{90.62}	&\darkblue{95.16}	&87.98\\
~ &GNNSAFE++   w/ $\mathcal{L}_{LN}$ 	&Yes	&90.13	&94.11	&\darkblue{93.83}		&\darkblue{84.93}	&\darkblue{87.68}	&91.00		&86.21	&87.56	&89.66\\

~ &NODESAFE++ (ours)	&Yes	&\darkred{94.64}	&\darkred{96.56}	&\darkred{94.88}		&\darkred{86.90}	&\darkred{91.14}	&\darkred{91.98}		&\darkred{96.30}	&\darkred{95.26}	&\darkred{93.48}\\

\hline
\hline

\multirow{14}{*}{\rotatebox{90}{AUPR ($\uparrow$)}} &MSP	&No	&45.73	&73.70	&78.03		&34.78	&55.48	&64.03		&17.44	&39.29	&34.98\\
~ &ODIN	&No	&27.01	&26.96	&24.27		&23.07	&23.11	&17.97		&4.83	&4.83	&13.49\\
~ &Mahalanobis	&No	&29.03	&31.95	&42.31		&21.20	&31.20	&35.47		&8.38	&15.09	&23.40\\
~ &Energy	&No	&46.08	&74.42	&78.14		&33.63	&55.94	&64.10		&17.32	&39.10	&36.00\\
~ &GKDE	&No	&44.26&	66.52	&27.50	&	31.55	&50.25&	61.21		&16.89	&32.41	&34.63\\
~ &GPN	&No	&53.26	&73.79	&77.40		&41.12	&53.21	&62.32		&17.54	&39.75	&35.12\\
~ &GNNSAFE	&No	&77.46	&\darkblue{88.19}	&82.21		&60.81	&67.02	&65.26		&\darkblue{62.74}	&\darkblue{71.66}	&44.77\\
~ &GNNSAFE w/  $\mathcal{L}_{LN}$	&No	&\darkblue{78.13}	&86.89	&\darkblue{85.19}		&\darkblue{69.73}	&\darkblue{76.20}	&\darkblue{67.69}		&58.72	&64.21	&\darkblue{54.33}\\
~ &NODESAFE (ours)	&No	&\darkred{83.98}	&\darkred{88.82}	&\darkred{85.22}		&\darkred{75.93}	&\darkred{79.30}	&\darkred{68.15}		&\darkred{71.29}	&\darkred{78.22}	&\darkred{71.98}\\

\cline{2-12}											

~ &OE	&Yes	&46.93	&70.84	&77.01		&30.07	&48.80	&62.74		&16.74	&38.60	&29.88\\
~ &Energy FT	&Yes	&49.18	&75.99	&78.49		&36.01	&55.69	&66.66		&18.00	&37.21	&\darkblue{52.39}\\
~ &GNNSAFE++	&Yes	&\darkblue{81.88}	&\darkblue{90.27}	&82.64		&65.58	&68.06	&65.48		&\darkblue{72.78}	&\darkblue{77.47}	&41.43\\
~ &GNNSAFE++   w/ $\mathcal{L}_{LN}$ 	&Yes	&81.61	&88.82	&\darkblue{84.17}		&\darkblue{70.90}	&\darkblue{76.10}	&\darkblue{67.57}		&57.25	&58.53	&44.87\\
~ &NODESAFE++ (ours)	&Yes	&\darkred{85.63}	&\darkred{91.96}	&\darkred{86.66}		&\darkred{71.41}	&\darkred{79.48}	&\darkred{68.97}		&\darkred{81.88}	&\darkred{78.12}	&\darkred{53.45}\\
\hline
\hline
\end{tabular}
}
\end{table*}

\begin{table*}[!t]
% \scriptsize
\centering
\caption{Out-of-distribution detection results are measured by AUROC ($\uparrow$) / AUPR ($\uparrow$) / FPR95 ($\downarrow$) on the \textit{Twitch} dataset, where nodes in different subgraphs are considered OOD data, and the \textit{Arxiv} dataset, where papers published after 2017 are treated as OOD data. The in-distribution testing accuracy is reported for calibration. Detailed results on each OOD dataset (i.e., subgraph or year) are presented in Appendix \ref{Appendix-Additional Experiment Results}. GPN reports an out-of-memory issue on \textit{Arxiv} with a 24GB GPU.}\label{tabel-Twitch_A}
\resizebox{\linewidth}{!}{
\begin{tabular}{c|c|cccc|cccc}
\hline
\hline
\multirow{2}*{Model}  &OOD &\multicolumn{4}{c|}{Twitch} &\multicolumn{4}{c}{Arxiv}\\
~ &Expo &AUROC(↑)	&AUPR(↑)	&FPR95(↓)	&ID ACC(↑)		&AUROC(↑)	&AUPR(↑)	&FPR95(↓)	&ID ACC(↑)\\
\hline
\hline
MSP	&No	&33.59	&49.14	&97.45	&68.72		&63.91	&75.85	&90.59	&53.78\\
ODIN	&No	&58.16	&72.12	&93.96	&70.79		&55.07	&68.85	&100.00	&51.39\\
Mahalanobis	&No	&55.68	&66.42	&90.13	&70.51		&56.92	&69.63	&94.24	&51.59\\
Energy	&No	&51.24	&60.81	&91.61	&70.40		&64.20	&75.78	&90.80	&53.36\\
GKDE	&No	&46.48	&62.11	&95.62	&67.44		&58.32	&72.62	&93.84	&50.76\\
GPN	&No	&51.73	&66.36	&95.51	&68.09		&-	&-	&-	&-\\
GNNSAFE	&No	&\darkblue{66.82}	&\darkblue{70.97}	&\darkblue{76.24}	&70.40		&71.06	&80.44	&87.01	&53.39\\
GNNSAFE w/ $\mathcal{L}_{LN}$ 	&No	&57.50	&68.27	&94.12	&67.10		&\darkblue{71.50}	&\darkblue{80.71}	&\darkblue{85.93}	 &46.34\\
NODESAFE (ours)	&No	&\darkred{89.99}	&\darkred{93.33}	&\darkred{47.00}	&71.79		 &\darkred{72.44}	&\darkred{81.51}	&\darkred{84.27}	&51.20\\
\hline
\hline
OE	&Yes	&55.72	&70.18	&95.07	&70.73		&69.80	&80.15	&85.16	&52.39\\
Energy FT	&Yes	&84.50	&88.04	&61.29	&70.52		&71.56	&80.47	&80.59	&53.26\\
GNNSAFE++	&Yes	&\darkblue{95.36}	&97.12	&\darkblue{33.57}	&70.18		&\darkblue{74.77}	&\darkblue{83.21}	&\darkblue{77.43}	&53.50\\
GNNSAFE++  w/ $\mathcal{L}_{LN}$ 	&Yes	&95.33	&\darkblue{97.39}	&33.81	&70.11		&72.21	&81.57	&85.49	&46.36\\
NODESAFE++ (ours)	&Yes	&\darkred{98.50}	&\darkred{99.18}	&\darkred{3.43}	&71.85		&\darkred{75.49}	&\darkred{83.71}	&\darkred{75.24}	&52.93\\
\hline
\hline
\end{tabular}
}
\end{table*}

\begin{table*}[!t]
% \scriptsize
\centering
\caption{Ablation experiment. OOD detection results measured by AUROC ($\uparrow$) / AUPR ($\uparrow$) / FPR95 ($\downarrow$) on \textit{Cora}, \textit{Twitch}, and \textit{Arxiv}, where the results presented by Cora are averaged across the three OOD generation methods of Structure Manipulation, Feature Interpolation, and Label Leave-out.}\label{table-Ablation}
\resizebox{\linewidth}{!}{
\begin{tabular}{c|c|ccc|ccc|ccc}
\hline
\hline
\multirow{2}*{Model}  &OOD &\multicolumn{3}{c|}{Cora} &\multicolumn{3}{c|}{Arxiv} &\multicolumn{3}{c}{Twitch}\\
~ &Expo &AUROC(↑)	&AUPR(↑)	&FPR95(↓)&AUROC(↑)	&AUPR(↑)	&FPR95(↓)		&AUROC(↑)	&AUPR(↑)	&FPR95(↓)	\\
\hline
\hline
NODESAFE	&No &\darkred{94.39}	&\darkred{86.01}	&\darkred{26.04}&\darkred{72.44}	&\darkred{81.51}	&\darkred{84.27}	&\darkred{89.99}	&\darkred{93.33}	&\darkred{47.00}		 \\
NODESAFE  w/o $\mathcal{L}_{uniform}$	&No &\darkblue{92.57}	&\darkblue{84.33}	&\darkblue{32.07}&\darkblue{71.86}	&\darkblue{81.12}	&\darkblue{85.35}	&\darkblue{86.23}	&\darkblue{91.14}	&\darkblue{63.54}	 \\
NODESAFE  w/o $\mathcal{L}_{bound}$	&No &69.25	&60.18	&77.55&61.54	&70.89	&89.76	&65.42	&70.75	&85.19 \\
\hline
\hline
NODESAFE++ 	&Yes &\darkred{95.36}	&\darkred{88.08}	&\darkred{20.20} &\darkred{75.49}	&\darkred{83.71}	&\darkred{75.24}		&\darkred{98.50}	&\darkred{99.18}	&\darkred{3.43}		\\
NODESAFE++  w/o $\mathcal{L}_{uniform}$	&Yes &\darkblue{93.16}	&\darkblue{86.47}	&\darkblue{26.43} &\darkblue{73.17}	&\darkblue{82.04}	&\darkblue{78.65}	&\darkblue{96.23}	&\darkblue{97.75}	&\darkblue{16.44}	 \\
NODESAFE++  w/o $\mathcal{L}_{bound}$	&Yes &81.86	&75.77	&50.21&63.02	&71.21	&89.15	&81.51	&82.76	&57.49  \\
\hline
\hline
\end{tabular}
}
\label{T-Ablation}
\end{table*}

\section{Experiment}\label{sec-Experiment}
This section outlines our experimental validation process. Section \ref{subsec-Setup} details the experimental dataset, metrics, and baseline methodologies. Section \ref{subsec-Results} showcases the primary experimental findings.
\subsection{Setup}\label{subsec-Setup}
\paragraph{Datasets and Splits}
We utilize five widely used real-world datasets in node classification tasks: \textit{Cora} \citep{kipf2016semi}, \textit{Citeseer} \citep{kipf2016semi}, \textit{Pubmed} \citep{kipf2016semi}, \textit{ogbn-Arxiv} \citep{hu2020open}, and \textit{Twitch-Explicit} \citep{wu2022handling}. Specifically, \textit{Twitch} is employed for the multi-graph scenario, where out-of-distribution samples originate from a different graph (or subgraph) disconnected from any nodes in the training set. The other datasets are used for the single graph scenario, where OOD test samples exist on the same graph as training samples but are not observed during the training process. We follow the partitioning standards of GNNSAFE \citep{wu2023energy} for dataset splits. For the \textit{Cora}, \textit{Citeseer}, and \textit{Pubmed} datasets, we employ three approaches to generate OOD data: Structure Manipulation, Feature Interpolation, and Label Leave-out. For the \textit{Twitch} dataset, OOD samples are derived from other graphs. We divide the samples based on temporal context for the \textit{Arxiv} dataset. More specific details are elaborated in the Appendix \ref{Appendix-Dataset Information}.
\paragraph{Metric}
We adhere to the common practice of using AUROC, AUPR, and FPR95 as evaluation metrics for OOD detection. The performance of ID data is measured by accuracy on test nodes. Further details are provided in the Appendix \ref{Appendix-Evaluation Metrics}.

\paragraph{Baselines}
We conduct a comparative analysis between two categories of methods for OOD detection. The first category comprises baseline models that primarily focus on OOD detection in the visual domain, where inputs are assumed to be i.i.d. sampled: MSP\citep{hendrycks2016baseline}, ODIN \citep{liang2017enhancing}, Mahalanobis \citep{lee2018simple}, OE \citep{hendrycks2018deep} (trained with OOD exposure), Energy, and Energy Fine-Tune \citep{liu2020energy}, which are also trained with OOD exposure. The second category consists of baseline models specifically designed for handling node-level OOD data on graphs. We compared three methods: GKDE\citep{zhao2020uncertainty}, GPN\citep{stadler2021graph} and the current sota method GNNSAFE \citep{wu2023energy}. In addition, we compare our NODESAFE with a similar work, LogitNorm \citep{wei2022mitigating}, which is implemented by normalizing the logit vector during training so that it has a constant norm, as follows:
\begin{equation}\label{equa-L_logitnorm}
    \mathcal{L}_{LN} = \mathcal{L}_{logit\_norm}(\mathbf{z}, y) = -\mathrm{log}\frac{e^{\mathrm{z}_y/(\tau\lVert \mathbf{z} \rVert)}}{\sum_{c=1}^C e^{\mathrm{z}_c/(\tau\lVert \mathbf{z} \rVert)}}
\end{equation}
where $\tau$ denotes the temperature coefficient.
% We explain the differences between our approach, NODESAFE, and LogitNorm through logit visualization experiments.

\begin{figure}[!t]
	% \centering
  \subfigure[w/ $\mathcal{L}_{CE}$]{
	% \begin{minipage}[b]{0.47\linewidth}
		\includegraphics[width=0.3\linewidth]{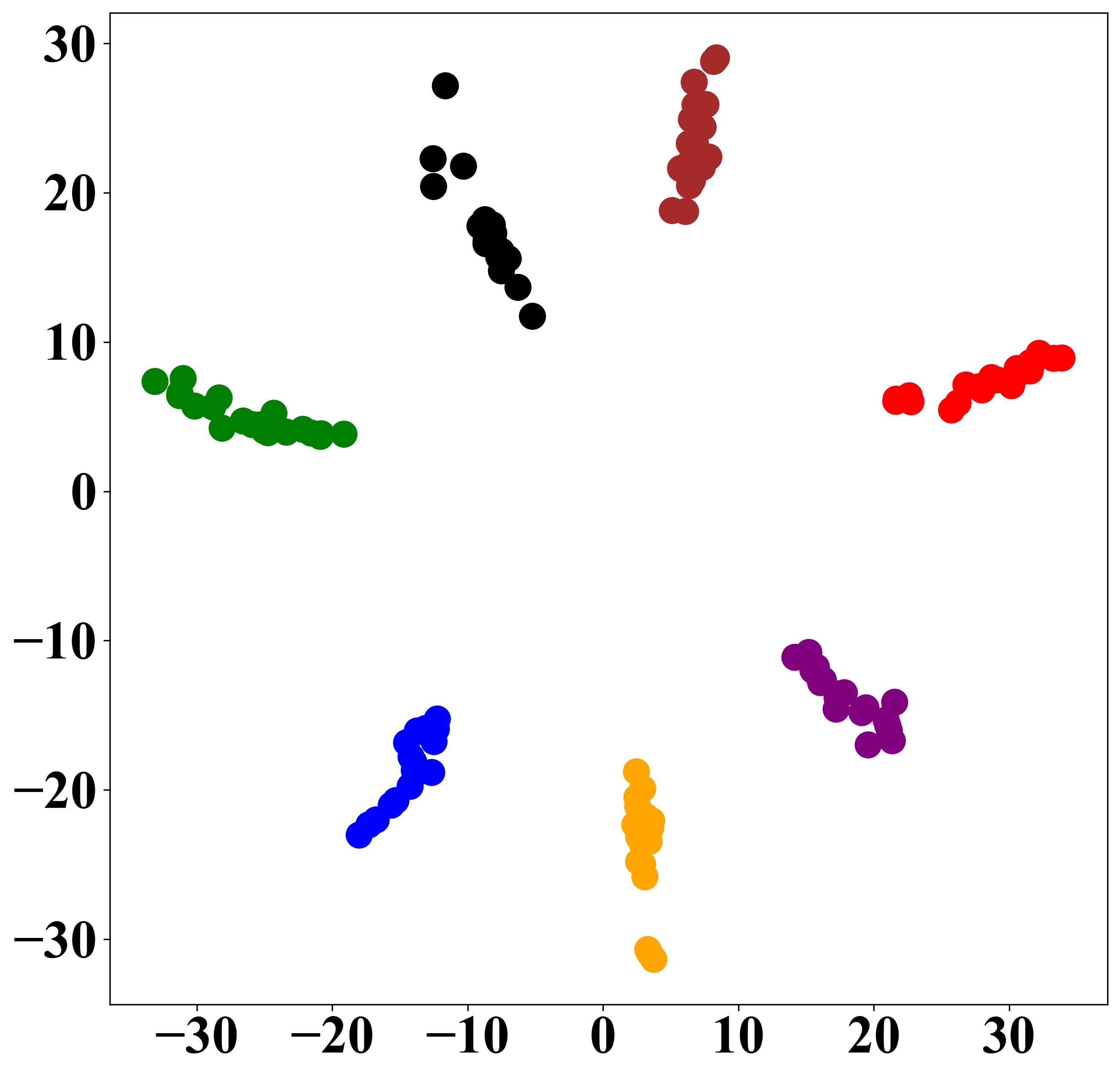}
	% \end{minipage}
 }
  \subfigure[w/ $\mathcal{L}_{CE} + \mathcal{L}_{\mathrm{LN}}$]{
    % \begin{minipage}[b]{0.47\linewidth}
		\includegraphics[width=0.3\linewidth]{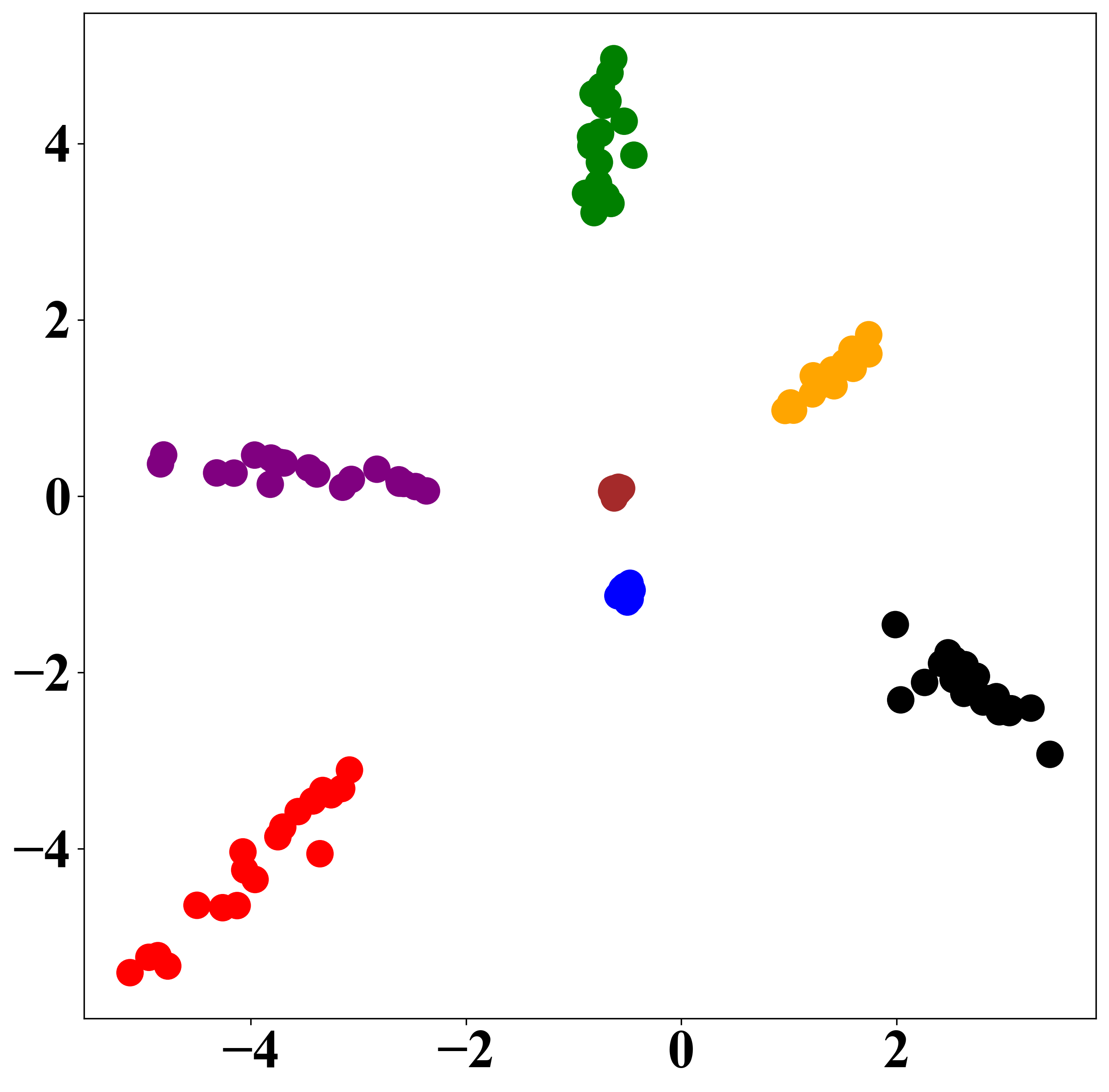}
	% \end{minipage}
 }
  \subfigure[w/ $\mathcal{L}_{CE} + \mathcal{L}_{\mathrm{UB}}$]{
    % \begin{minipage}[b]{0.47\linewidth}
		\includegraphics[width=0.3\linewidth]{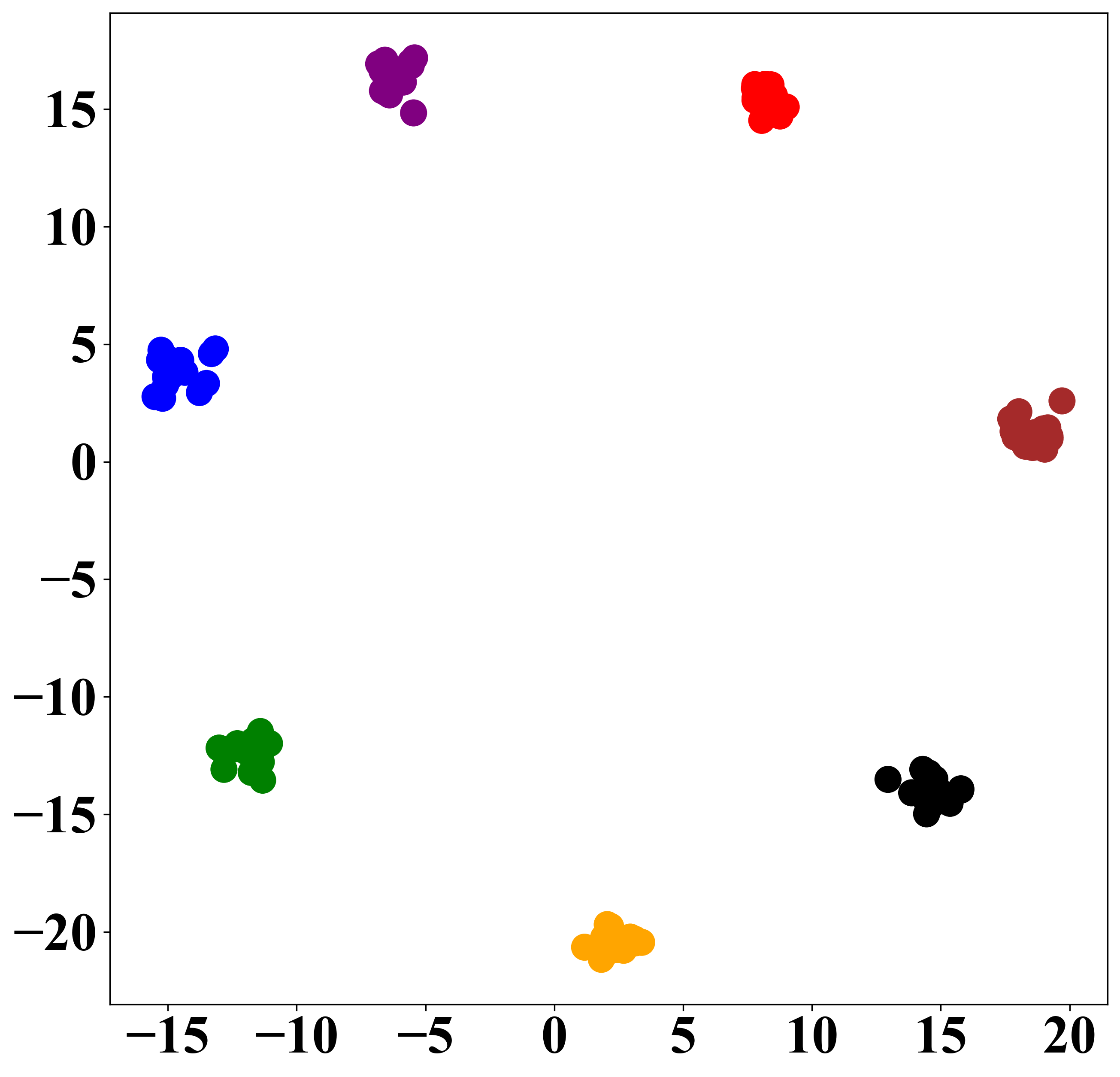}
	% \end{minipage}
 }
    \caption{Visualizing 2D logits of nodes using GAT for node classification on the \textit{Cora} dataset. The 2D logits are derived by splitting a single classification linear layer, $\mathbb{R}^D \rightarrow \mathbb{R}^C$, into two consecutive linear layers, $\mathbb{R}^D \rightarrow \mathbb{R}^2 \rightarrow \mathbb{R}^C$. Here, $D$ represents the node representation dimension, $C$ is the number of classification categories, and we visualize the logit in $\mathbb{R}^2$ space.}
    \label{F-logit_visual}
\end{figure}

\subsection{Results}\label{subsec-Results}
\paragraph{How does $\mathcal{L}_{\mathrm{UB}}$ affect the detection performance of OOD samples from different sources?}

\begin{figure*}[!t]
	% \centering
  \subfigure[Energy]{
	% \begin{minipage}[b]{0.47\linewidth}
		\includegraphics[width=0.23\linewidth]{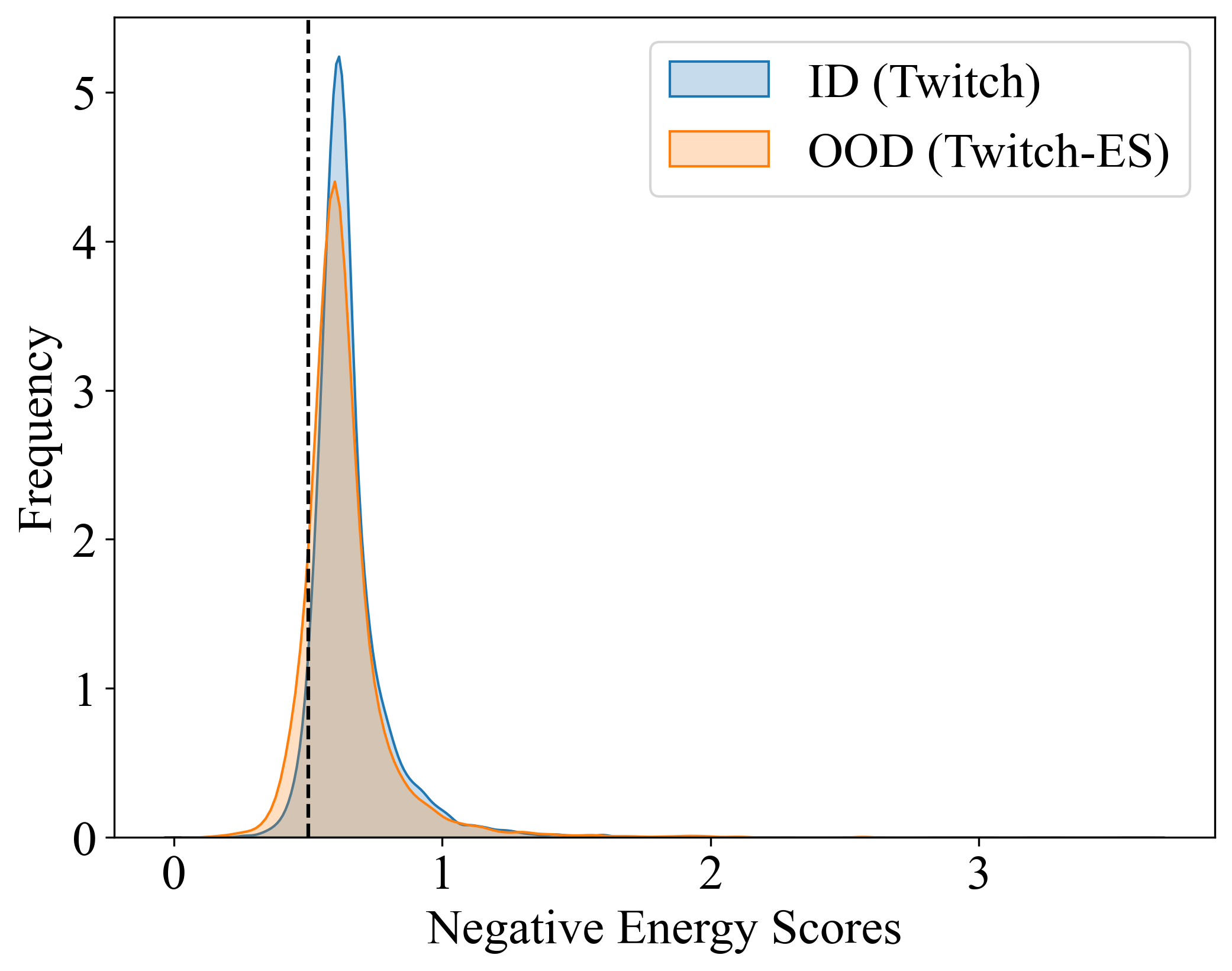}
	% \end{minipage}
 }
  \subfigure[GNNSAFE]{
    % \begin{minipage}[b]{0.47\linewidth}
		\includegraphics[width=0.23\linewidth]{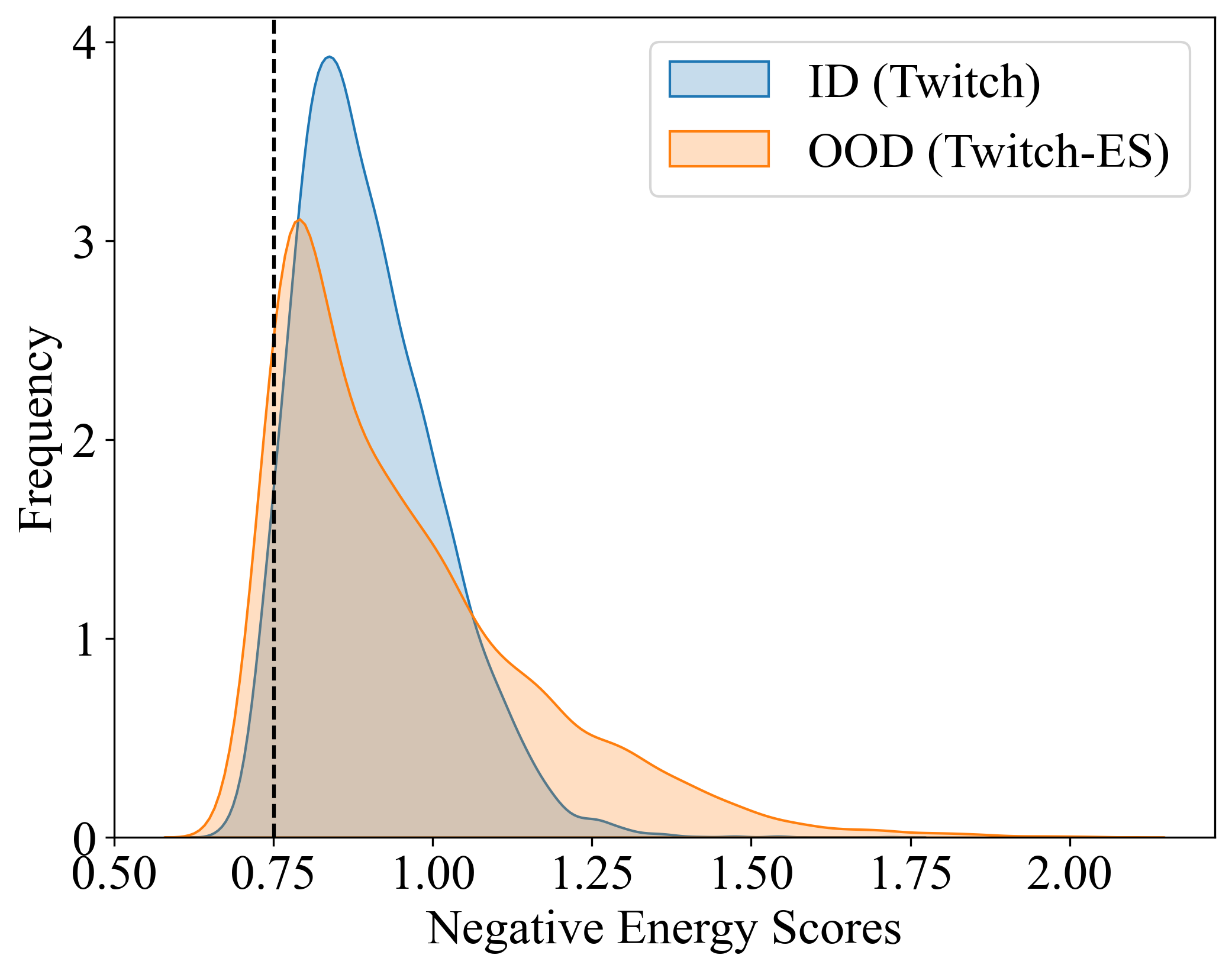}
	% \end{minipage}
 }
  \subfigure[GNNSAFE w/ $\mathcal{L}_{LN}$]{
    % \begin{minipage}[b]{0.47\linewidth}
		\includegraphics[width=0.23\linewidth]{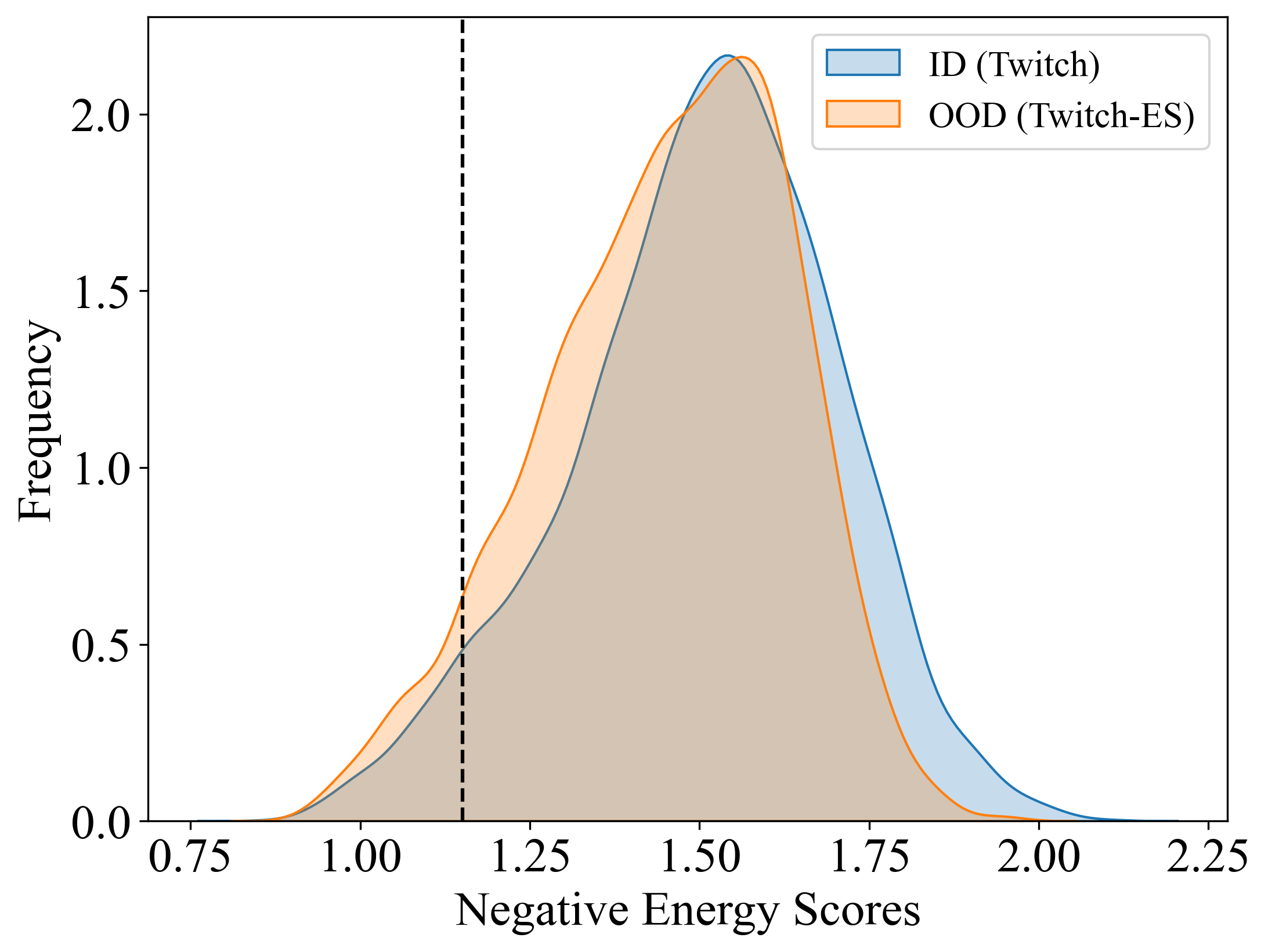}
	% \end{minipage}
 }
   \subfigure[NODESAFE (ours)]{
    % \begin{minipage}[b]{0.47\linewidth}
		\includegraphics[width=0.23\linewidth]{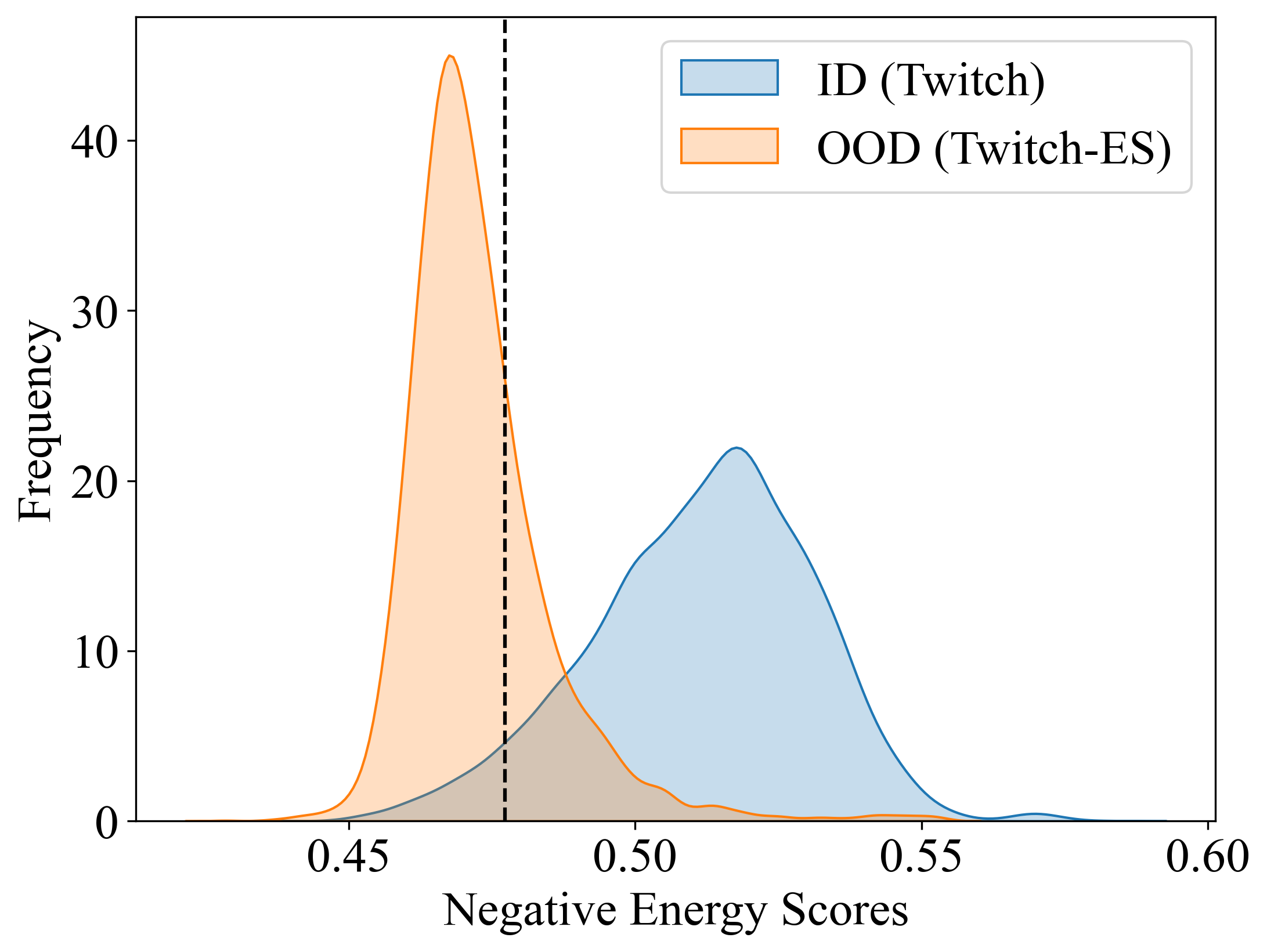}
	% \end{minipage}
 }
 
   \subfigure[Energy-FT]{
	% \begin{minipage}[b]{0.47\linewidth}
		\includegraphics[width=0.23\linewidth]{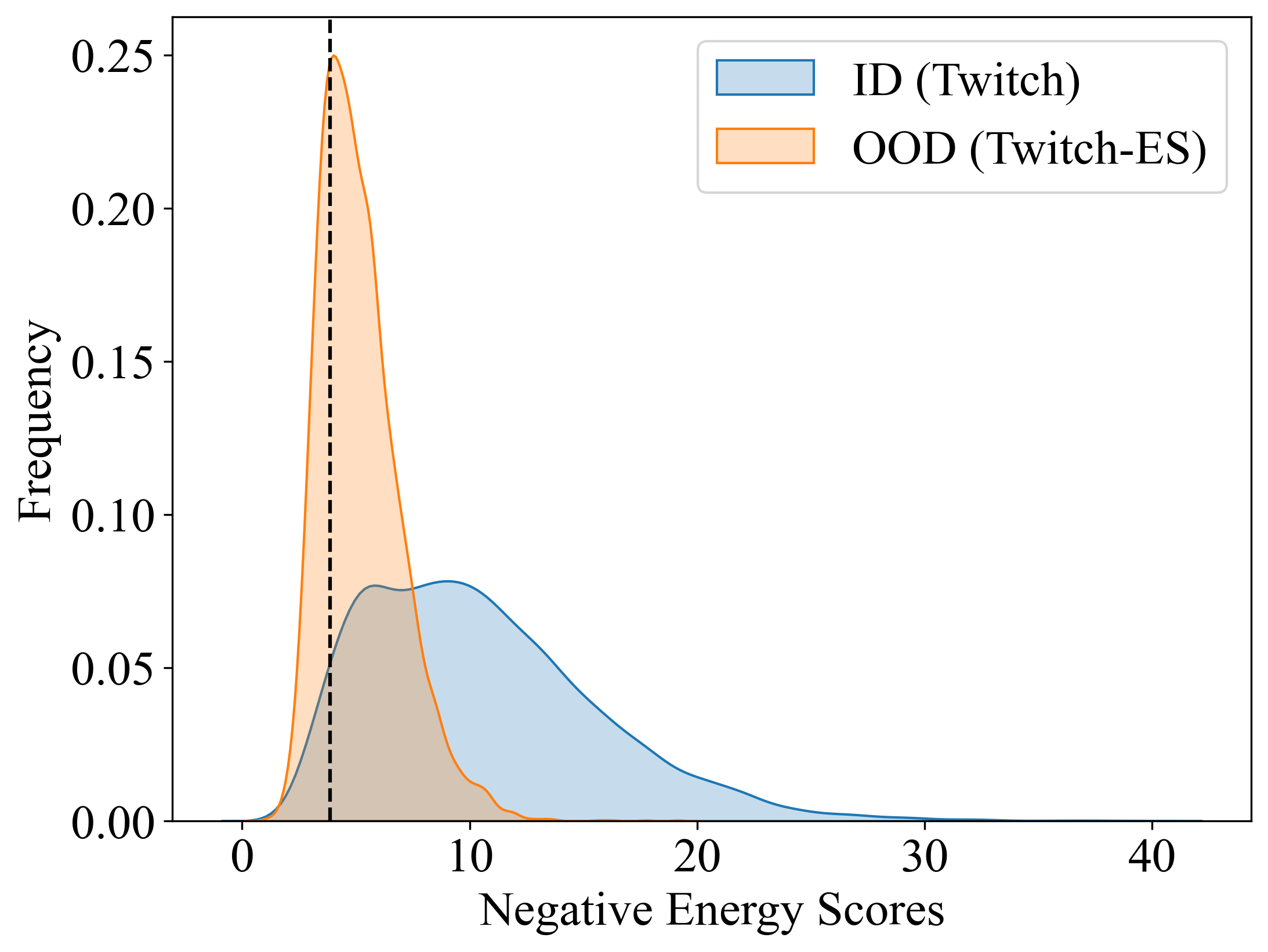}
	% \end{minipage}
 }
  \subfigure[GNNSAFE++]{
    % \begin{minipage}[b]{0.47\linewidth}
    \includegraphics[width=0.23\linewidth]{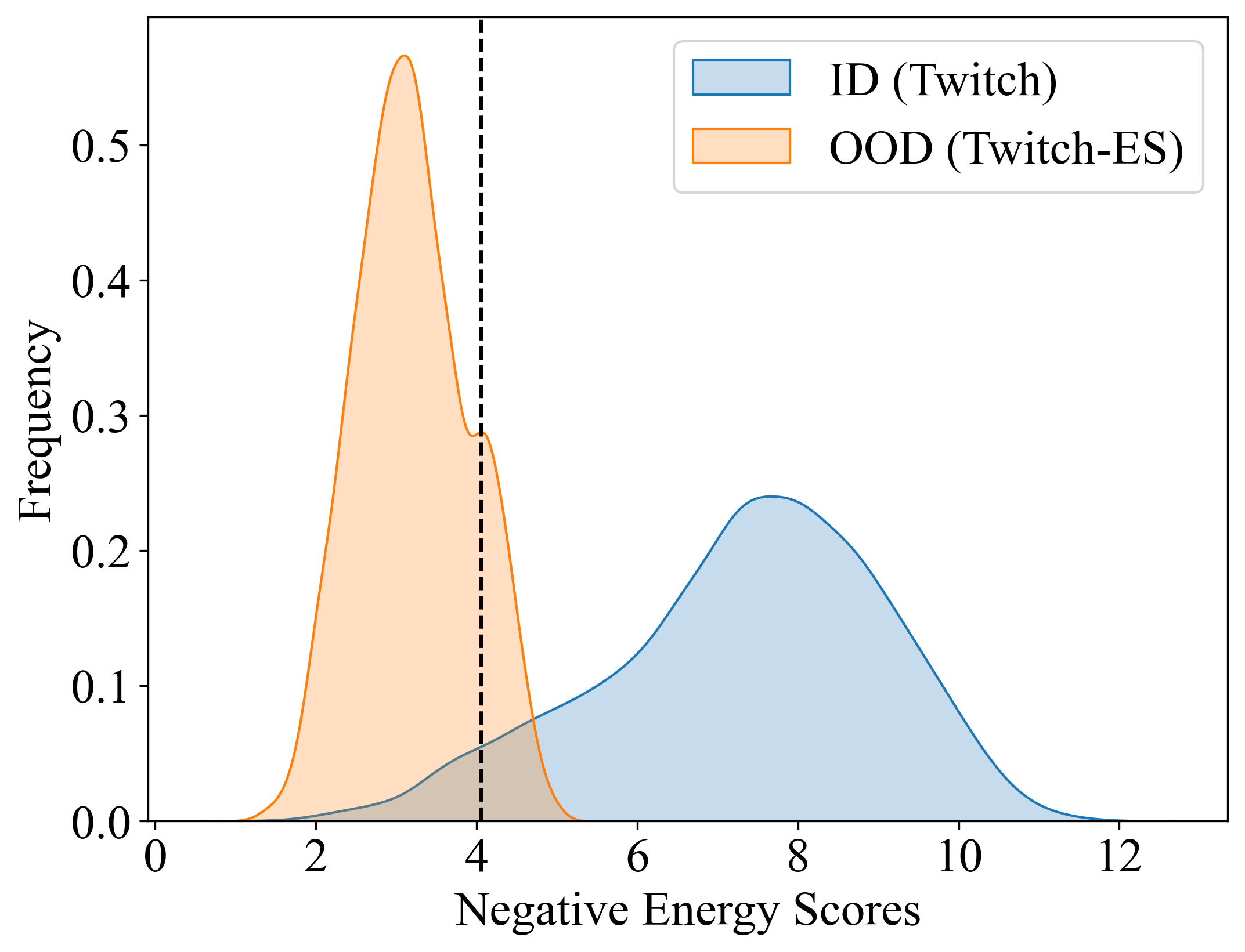}
	% \end{minipage}
 }
  \subfigure[GNNSAFE++ w/ $\mathcal{L}_{LN}$]{
    % \begin{minipage}[b]{0.47\linewidth}
		\includegraphics[width=0.23\linewidth]{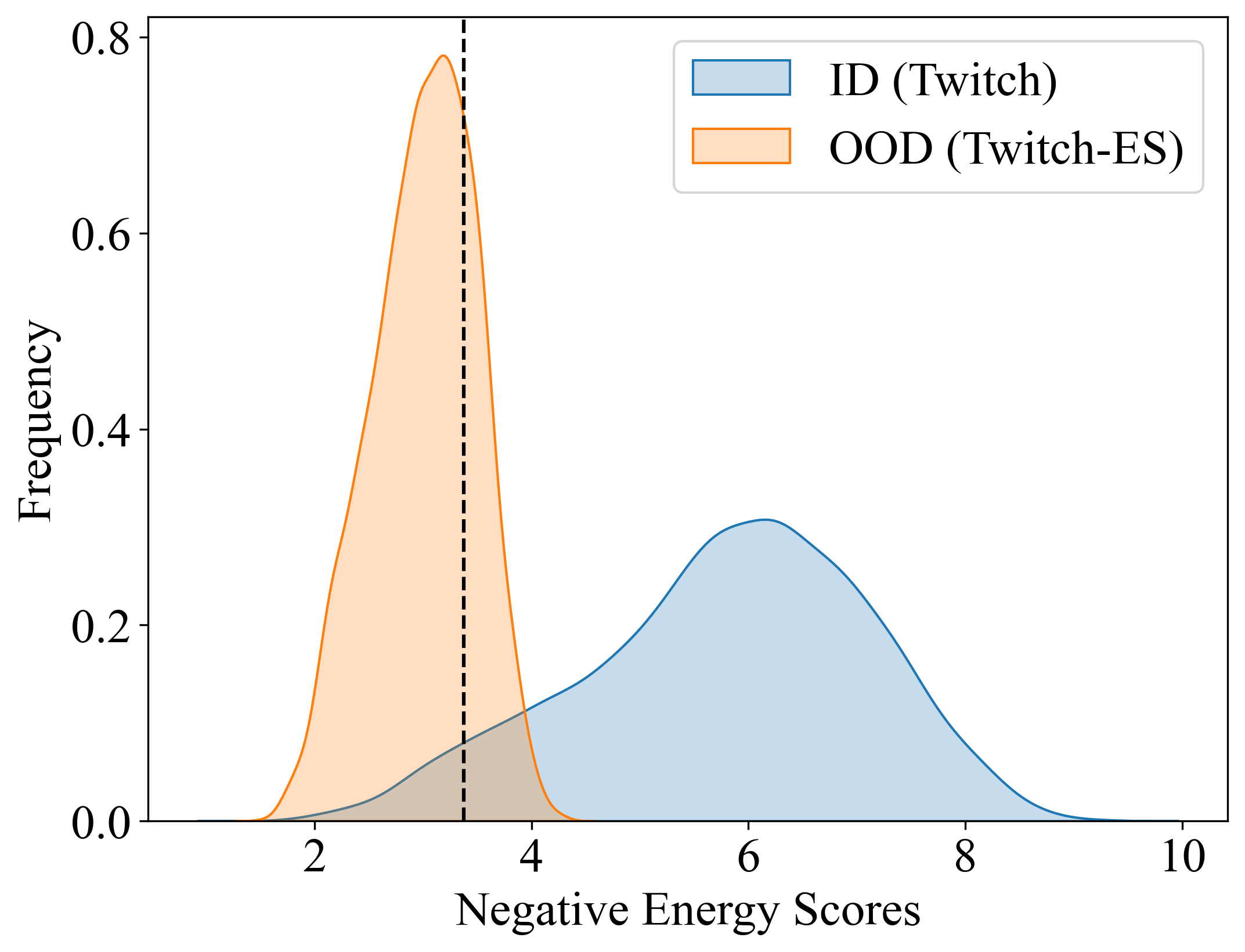}
	% \end{minipage}
 }
   \subfigure[NODESAFE++ (ours)]{
    % \begin{minipage}[b]{0.47\linewidth}
		\includegraphics[width=0.23\linewidth]{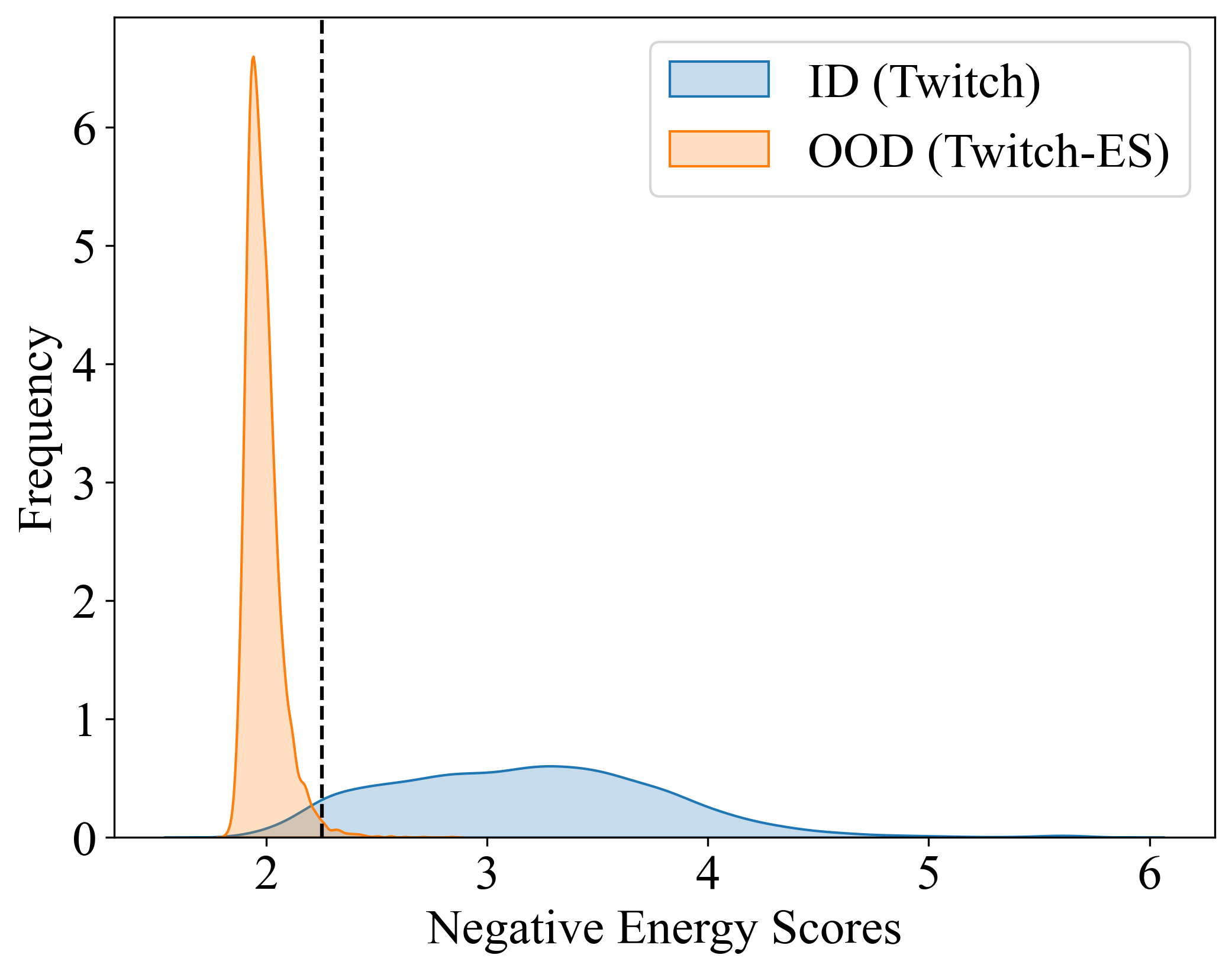}
	% \end{minipage}
 }
    \caption{The Negative Energy Score distributions on \textit{Twitch} where nodes in different sub-graphs are OOD samples.}
    \label{F-Frequency}
\end{figure*}

\begin{figure*}[!t]
	% \centering
  \subfigure[GNNSAFE ( ID )]{
	% \begin{minipage}[b]{0.47\linewidth}
		\includegraphics[width=0.23\linewidth]{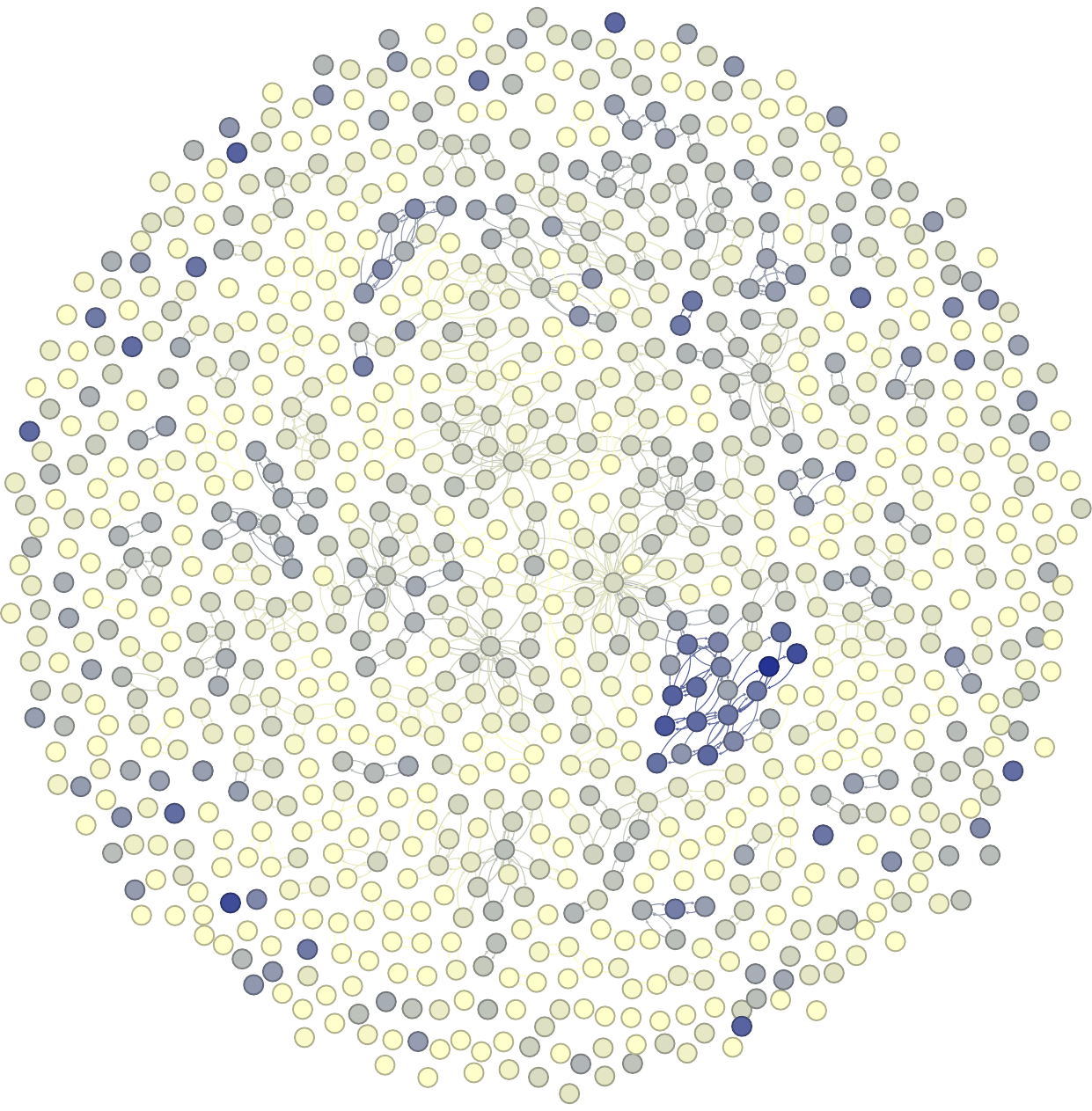}
	% \end{minipage}
 }
  \subfigure[NODESAFE ( ID )]{
    % \begin{minipage}[b]{0.47\linewidth}
		\includegraphics[width=0.23\linewidth]{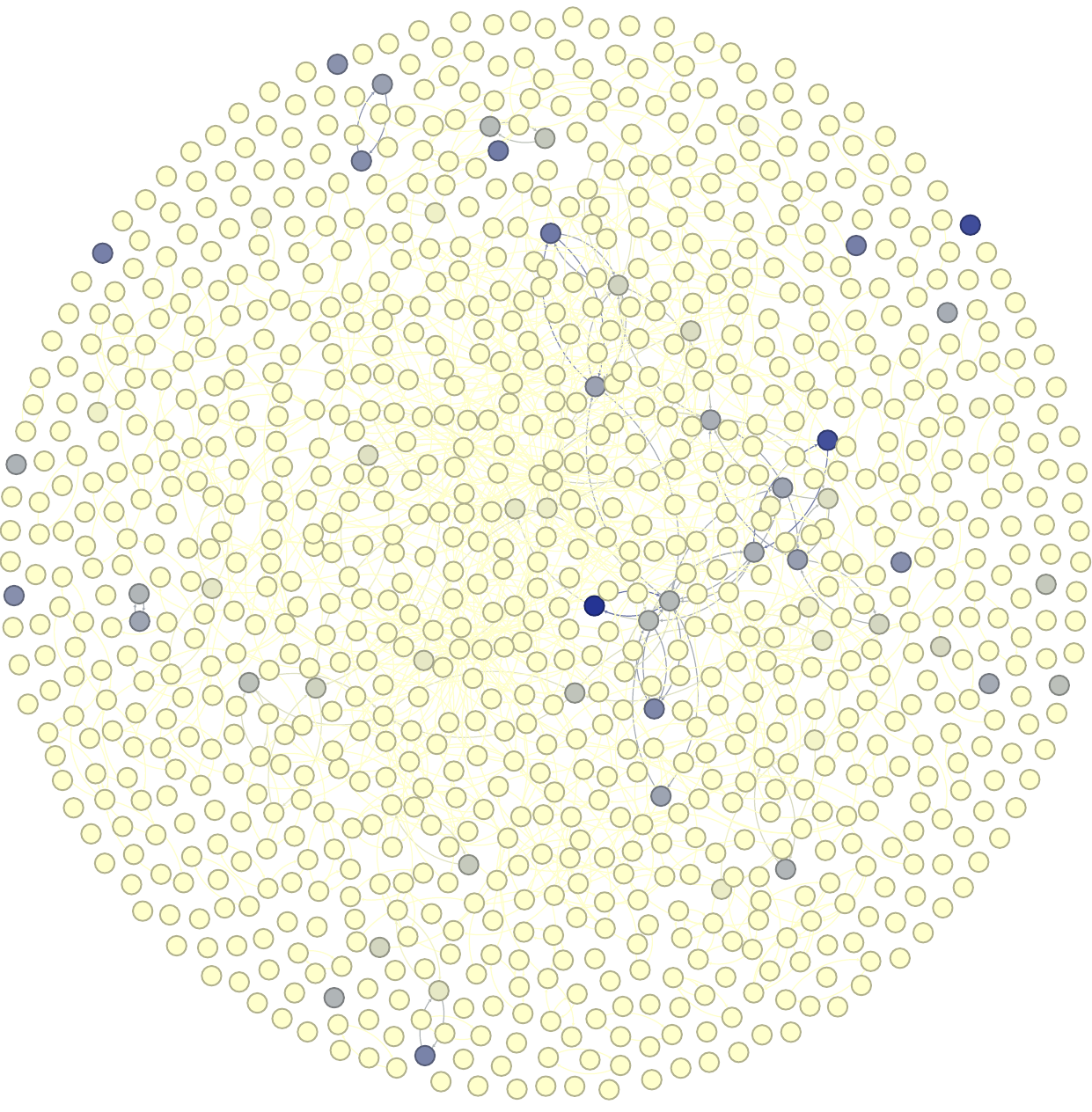}
	% \end{minipage}
 }
  \subfigure[GNNSAFE ( OOD )]{
    % \begin{minipage}[b]{0.47\linewidth}
		\includegraphics[width=0.23\linewidth]{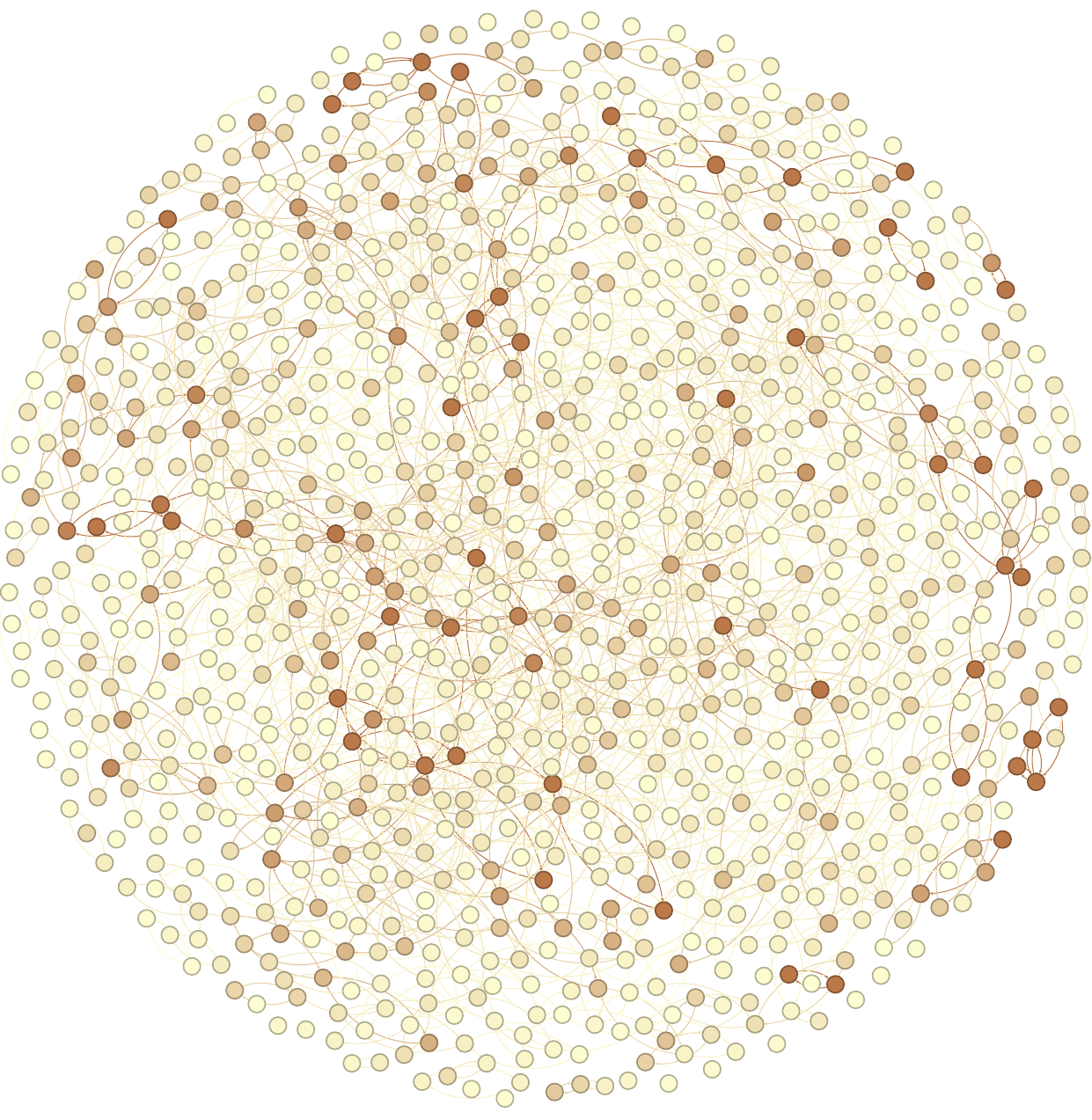}
	% \end{minipage}
 }
   \subfigure[NODESAFE ( OOD )]{
    % \begin{minipage}[b]{0.47\linewidth}
		\includegraphics[width=0.23\linewidth]{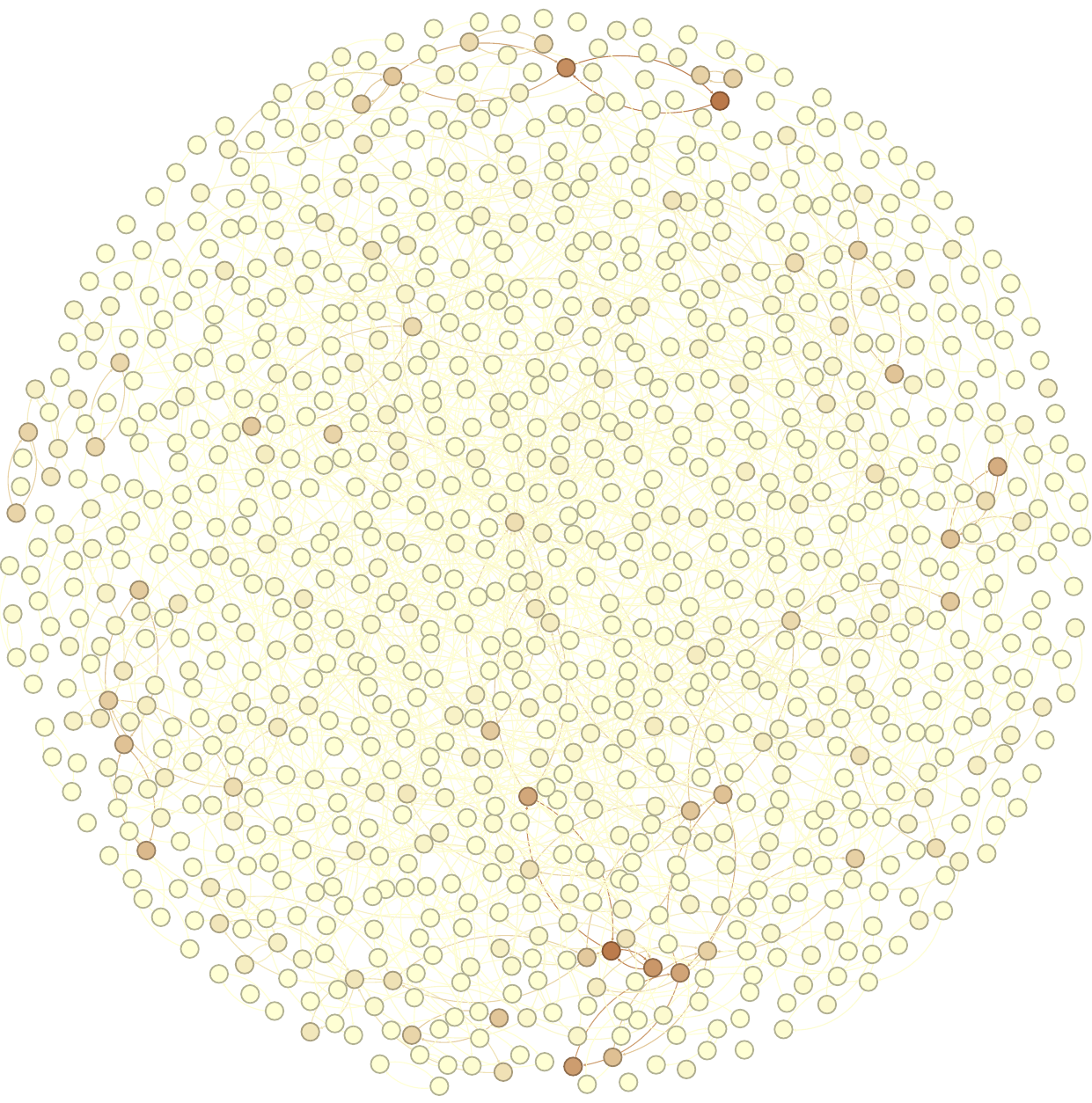}
	% \end{minipage}
 }
    \caption{Visualization of negative energy scores for different nodes. We compare the distribution of scores of GNNSAFE and NODESAFE on ID and OOD nodes after score aggregation on the \textit{Cora} dataset.}
    \label{F-Energy-global}
\end{figure*}

\begin{figure}[!t]
	\centering
  \subfigure[Epoch]{
	% \begin{minipage}[b]{0.47\linewidth}
		\includegraphics[width=0.3\linewidth]{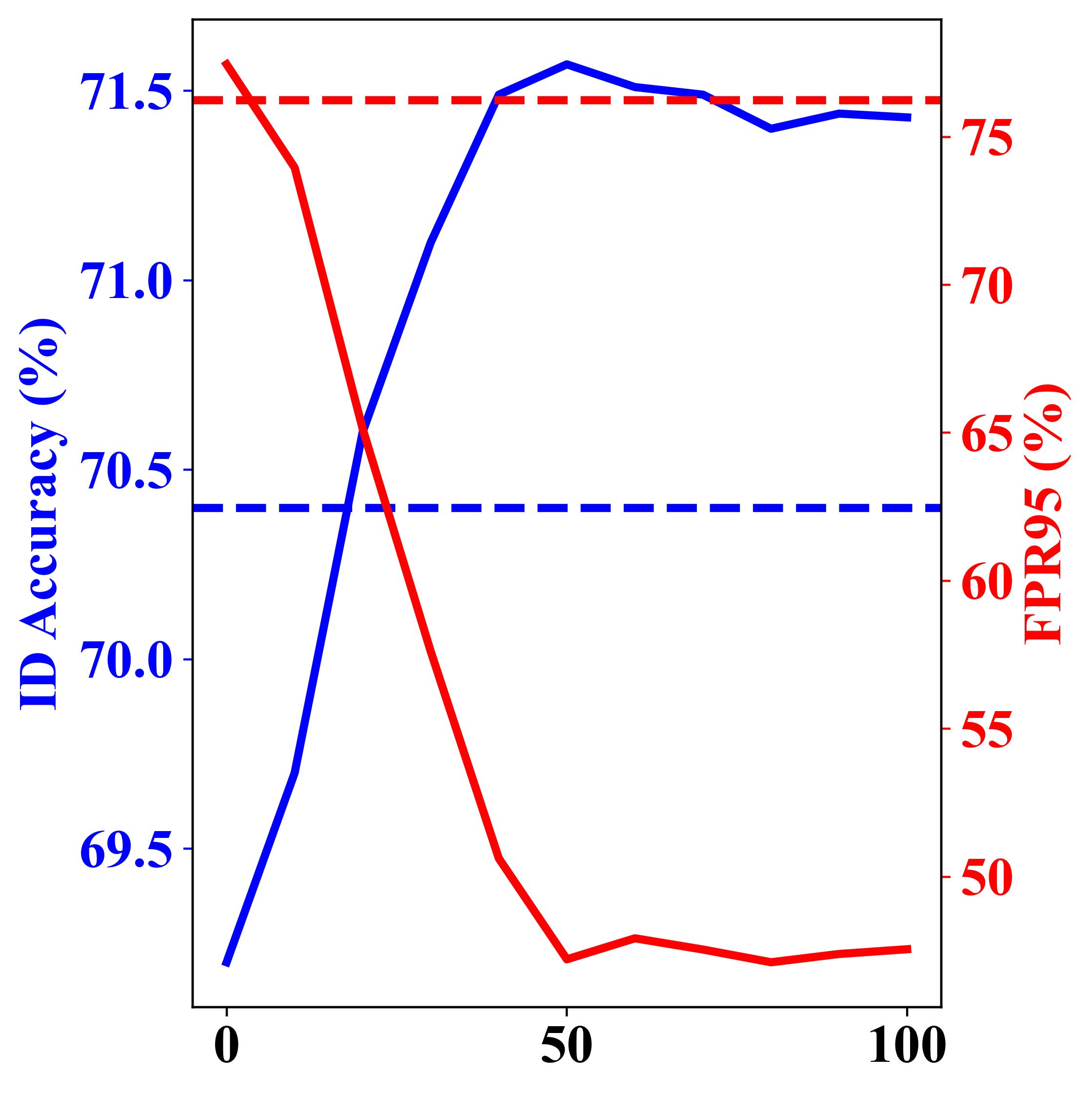}
	% \end{minipage}
 }
  \subfigure[$\lambda_1$]{
    % \begin{minipage}[b]{0.47\linewidth}
		\includegraphics[width=0.3\linewidth]{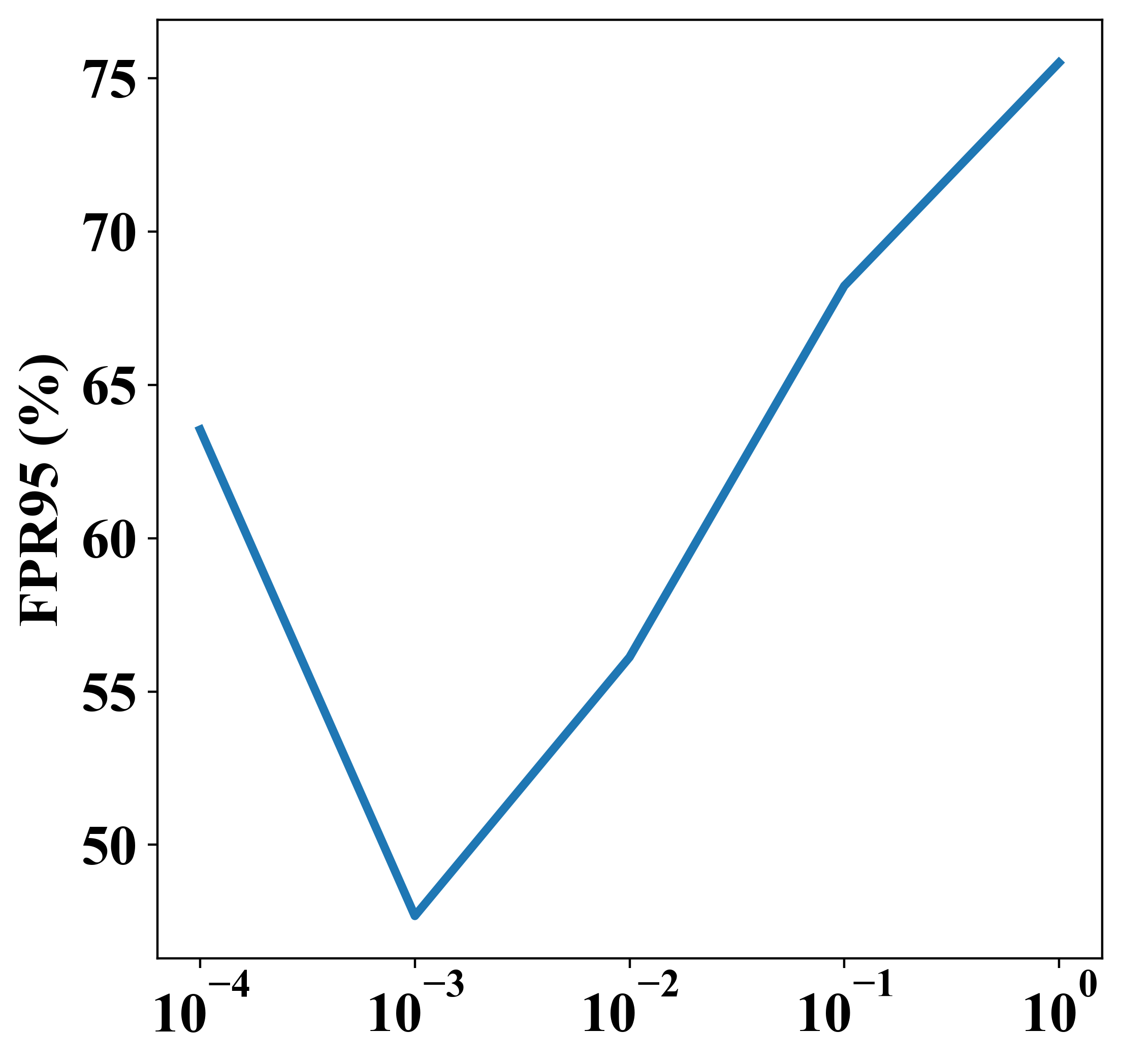}
	% \end{minipage}
 }
   \subfigure[$\lambda_2$]{
    % \begin{minipage}[b]{0.47\linewidth}
		\includegraphics[width=0.3\linewidth]{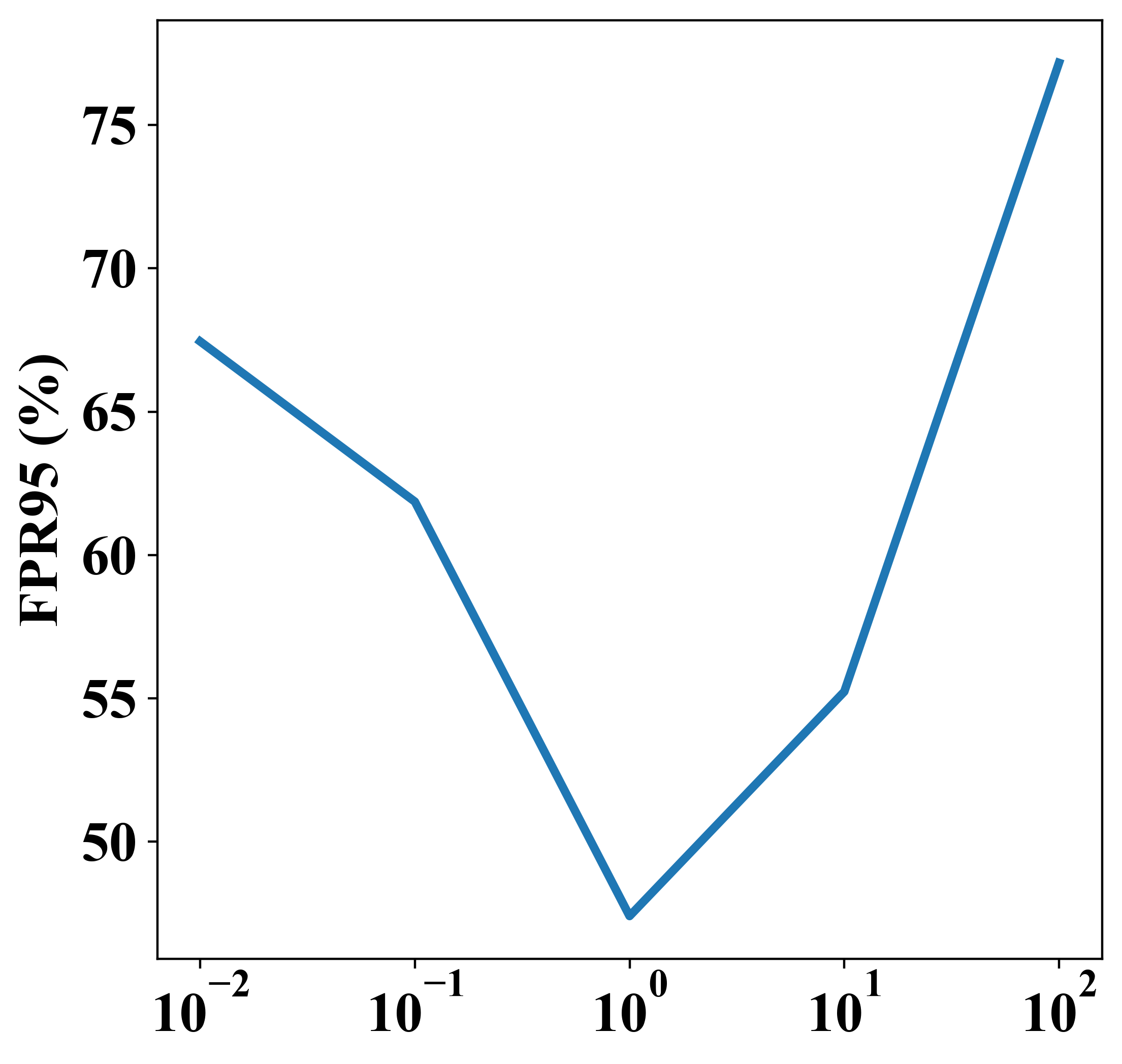}
	% \end{minipage}
 }
    \caption{Hyperparameter analysis on the \textit{Twitch} dataset.}
    \label{F-Hyperparameter}
\end{figure}

The introduction of the $\mathcal{L}_{\mathrm{UB}}$ optimization term results in a significant improvement in the model's detection capability. For instance, Tables \ref{tabel-Cora_C_P} and \ref{tabel-Twitch_A} illustrate the FPR95 metric for OOD data originating from Structure Manipulation. When $\mathcal{L}_{\mathrm{UB}}$ is incorporated, in scenarios without (with) OOD data exposure, it leads to an average FPR95 reduction of \textbf{28.4}\% (\textbf{22.7}\%). Similarly, for OOD data from Feature Interpolation, $\mathcal{L}_{\mathrm{UB}}$ achieves an average reduction of 18\% (15.8\%). In contrast, Label Leave-out achieves an average decrease of 6.8\% (5.8\%) in scenarios without (with) OOD data exposure. For time-based OOD data, $\mathcal{L}_{\mathrm{UB}}$ results in an average decrease of 2.74\% (2.19\%) in scenarios without (with) OOD data exposure. In multi-graph datasets with OOD data originating from other graphs, $\mathcal{L}_{\mathrm{UB}}$ leads to an average reduction of \textbf{29.2}\% (\textbf{30.1}\%) in scenarios without (with) OOD data exposure.

Remarkably, $\mathcal{L}_{\mathrm{UB}}$ has demonstrated substantial performance improvement in model detection for OOD samples in scenarios without OOD data exposure, addressing challenges associated with the limited availability of OOD data in real-world scenarios. The performance improvement of $\mathcal{L}_{\mathrm{UB}}$ for OOD data based on the temporal split is more limited compared to Structure Manipulation, Feature Interpolation, and Label Leave-out. We believe there are two possible reasons for this: 1. The closeness of the negative energy scores between the ID data and the OOD data, in this case, leads to similar means. 2. The performance of supervised classification on ID data is not good enough; hence, the generated logits are more confusing. Therefore, $\mathcal{L}_{\mathrm{UB}}$ brings limited improvement in OOD detection performance. Our future research work will aim to make further improvements.

\paragraph{What is the difference between our  method $\mathcal{L}_{\mathrm{UB}}$ and LogitNorm $\mathcal{L}_{\mathrm{LN}}$?}
Logitnorm is implemented by normalizing the logit vector to have a constant norm during training. Since LogitNorm only uses mathematical normalization to convert the logit, the logit formed by it still has significant variance on 2-norm, such as in Fig. \ref{F-logit_visual}. As a result, LogitNorm does not bring as effective a boost to OOD detection at the node level as our approach NODESAFE, as verified by the experimental results in Tables \ref{tabel-Cora_C_P} and \ref{tabel-Twitch_A}.
\paragraph{What are the effects of $\mathcal{L}_{\mathrm{uniform}}$ and $\mathcal{L}_{\mathrm{bound}}$ on OOD detection, respectively?}

We conduct ablation experiments on the \textit{Cora}, \textit{Arxiv}, and \textit{Twitch} datasets. From Table \ref{T-Ablation}, We observe that the direct use of $\mathcal{L}_{\mathrm{uniform}}$ does not significantly improve the out-of-distribution detection of the model. We speculate that this may be due to the lack of upper and lower bound constraints on the negative energy score, making it challenging to achieve robust control of the logit shift. Conversely, the effect of $\mathcal{L}_{\mathrm{bound}}$ alone already provides a notable enhancement, indicating that the unboundedness of the negative energy scores is the root cause of the excessive variance of the aggregated scores. Therefore, we conclude that the combination of $\mathcal{L}_{\mathrm{uniform}}$ and $\mathcal{L}_{\mathrm{bound}}$ is more effective in improving the performance of OOD detection. 
%However, it is essential to establish the use of $\mathcal{L}_{\mathrm{uniform}}$ on top of $\mathcal{L}_{\mathrm{bound}}$.

\paragraph{Visualization of different methods.}
Fig. \ref{F-Frequency} shows the frequency density plots of the negative energy scores of the samples obtained by the different methods on the \textit{Twitch} dataset, where the \textit{Twitch-ES} graph is used as OOD data. We find that the negative energy scores of the $\mathcal{L}_{\mathrm{UB}}$ order ID data and the OOD data were significantly closer to the mean, which led to a good classification of the sample points that also deviated from the mean. In scenarios with OOD data exposure, the combined effect of $\mathcal{L}_{\mathrm{UB}}$ and hinge loss $\mathcal{L}_{reg}$ makes it more likely to detect OOD data successfully. Fig. \ref{F-Energy-global} shows the distribution of GNNSAFE and NODESAFE negative energy scores after aggregation. The results demonstrate that without $\mathcal{L}_{\mathrm{UB}}$, the variance of the negative energy scores for both ID and OOD is higher, and the extreme scores significantly affect the scores of neighboring nodes. After adding the $\mathcal{L}_{\mathrm{UB}}$ constraint, the number of nodes with extreme scores is significantly reduced, and the overall of each of the ID and OOD nodes becomes more uniform.

\paragraph{Hyperparameter analysis.}
 Fig. \ref{F-Hyperparameter}(a) shows the relationship between $\mathcal{L}_{\mathrm{UB}}$ addition at different epochs and the final ID accuracy and FPR95 (solid line), with the dashed line indicating the GNNSAFE benchmark. We find $\mathcal{L}_{\mathrm{UB}}$ works better when the model is good at the supervised classification task. Because the logit at the beginning of training is chaotic, adding $\mathcal{L}_{\mathrm{UB}}$ in this case not only cannot enhance the OOD detection ability of the model but also affects the supervised classification performance of the model. Fig. \ref{F-Hyperparameter}(b) shows that performance is improved in both $\mathcal{L}_{\mathrm{uniform}}$ and $\mathcal{L}_{\mathrm{bound}}$ ratios between $10^{-4}$ and $10^{-2}$ orders of magnitude, with the best results at $10^{-3}$. Fig. \ref{F-Hyperparameter}(c) shows that the performance is improved in the ratio of $\mathcal{L}_{\mathrm{OOD}}$ and $\mathcal{L}_{\mathrm{UB}}$ between $10^{-1}$ and $10^{1}$ orders of magnitude, with the best results at $10^{0}$.

\section{Related Work}\label{sec-related_work}
\paragraph{Graph-level out-of-distribution detection}
Some OOD detection methods proposed in the vision domain can be directly used for graph-level OOD detection, which can be roughly divided into two groups. One group aims to design scoring functions that can be directly used for graph-level OOD detection, such as OpenMax score \citep{bendale2016towards}, MSP\citep{hendrycks2016baseline}, ODIN \citep{liang2017enhancing}, Mahalanobis \citep{lee2018simple}, Energy, and Energy Fine-Tune \citep{liu2020energy}, ReAct \citep{sun2021react}, GradNorm score \citep{huang2021importance}, and non-parametric KNN-based score \citep{sun2022out,zhu2022detecting}.
Another group addresses the out-of-distribution detection problem by training-time regularization\citep{lee2018simple,bevandic2018discriminative,hendrycks2018deep,geifman2019selectivenet,malinin2018predictive,mohseni2020self,jeong2020ood,chen2021atom,wei2021open,wei2022mitigating,ming2022cider,ming2022poem}. In addition, there has been some work on graph-domain-specific methods. A recent study \citep{li2022graphde} models the graph generation process by defining a variational distribution used to infer the environment. Furthermore, \cite{bazhenov2022towards} considers the graph OOD detection problem from the uncertainty estimation perspective and finds that it is essential to consider both graph representations and predictive classification distributions. GOOD-D \citep{liu2023good} develops a novel graph contrastive learning framework for detecting graphs without relying on ground truth labels. Using data-centric operations, \citep{guo2023data} designs a graph OOD detection adaptive amplifier post-framework. \citep{ding2023sgood} exploits substructures to learn powerful representations for OOD detection. These graph-level works still study different graphs as independent samples and cannot be directly used for node-level OOD detection.

\paragraph{Node-level out-of-distribution detection}
% By collecting evidence from the given labels of training nodes, \citep{zhao2020uncertainty} designs a graph-based kernel Dirichlet distribution estimation (GKDE) method for accurately predicting node-level Dirichlet distributions and detecting OOD nodes. \citep{stadler2021graph} proposes a new model, Graph Posterior Networks (GPNs), which explicitly performs Bayesian a posteriori updating for predicting interdependent nodes. Inspired by the label propagation algorithm, \citep{wu2023energy} proposes on-graph OOD detection based on aggregation of negative energy scores. Compared to the work on OOD detection at the graph level, less work has been done on OOD detection at the node level, which is an urgent research problem.

By collecting evidence from the given labels of training nodes, \citep{zhao2020uncertainty} designs a graph-based kernel Dirichlet distribution estimation (GKDE) method for accurately predicting node-level Dirichlet distributions and detecting out-of-distribution (OOD) nodes. \citep{stadler2021graph} proposes a new model, Graph Posterior Networks (GPNs), which explicitly performs Bayesian a posteriori updating for predicting interdependent nodes. Inspired by the label propagation algorithm, \citep{wu2023energy} proposes on-graph OOD detection based on aggregating negative energy scores. Compared to the work on OOD detection at the graph level, less effort has been devoted to OOD detection at the node level, which remains an urgent research problem.

\section{Conclusion}\label{sec-conclusion}
We propose a node-level graph OOD detection based on bounded and uniform negative energy score aggregation. Specifically, we find that graph OOD detection based on negative energy score aggregation suffers from extreme score variance in both ID and OOD samples. This phenomenon arises for two reasons: one is due to the negative energy scores being unbounded, and the other is due to logit shifts. Logit shifts are the direct cause of the excessive score variance. In contrast, the unboundedness of the negative energy scores is the root cause, which provides the conditions for generating extreme scores in the ID and the OOD samples. For these two causes, we make the negative energy scores bounded as well as mitigate the logit shift through two optimization objectives $\mathcal{L}_\mathrm{bound}$ and $\mathcal{L}_\mathrm{uniform}$ that bound the variance, respectively. Experimental results show that our approach significantly improves the node-level OOD detection performance.

\section*{Acknowledgement}
This work is partially supported by the National Natural Science Foundation of China (62172202), Collaborative Innovation Center of Novel Software Technology and Industrialization, the Major Program of the Natural Science Foundation of Jiangsu Higher Education Institutions of China under Grant No.22KJA520008, and the Priority Academic Program Development of Jiangsu Higher Education Institutions.

\section*{Impact Statement}
This paper presents work whose goal is to advance the field of Machine Learning. There are many potential societal consequences of our work, none of which we feel must be specifically highlighted here.

%%%%%%%%%%%%%%%%%%%%%%%%%%%%%%%%%%%%%%%%%%%%%%%%%%%%%%%%%%%%%%%

% In the unusual situation where you want a paper to appear in the
% references without citing it in the main text, use \nocite
\nocite{langley00}

\bibliography{ICML2024-main}
\bibliographystyle{icml2024}

%%%%%%%%%%%%%%%%%%%%%%%%%%%%%%%%%%%%%%%%%%%%%%%%%%%%%%%%%%%%%%%%%%%%%%%%%%%%%%%
%%%%%%%%%%%%%%%%%%%%%%%%%%%%%%%%%%%%%%%%%%%%%%%%%%%%%%%%%%%%%%%%%%%%%%%%%%%%%%%
% APPENDIX
%%%%%%%%%%%%%%%%%%%%%%%%%%%%%%%%%%%%%%%%%%%%%%%%%%%%%%%%%%%%%%%%%%%%%%%%%%%%%%%
%%%%%%%%%%%%%%%%%%%%%%%%%%%%%%%%%%%%%%%%%%%%%%%%%%%%%%%%%%%%%%%%%%%%%%%%%%%%%%%
\newpage
\appendix
\onecolumn

\section{Proof of Technical Results}\label{Appendix-sec-proof}

\subsection{Proof of Proposition \ref{prop-logitshift}}\label{Appendix-Proof-Logitshift}
\begin{proof}
Consider the softmax function denoted by $\sigma$ for the softmax cross-entropy loss. Given any constant $s$, if $\mathrm{c=argmax_i(\mathrm{z}_i)}$, then
\[
    \sigma(\mathrm{z}_c + s) 
    = \frac{exp(\mathrm{z}_c+s)}{\sum_{j=1}^D exp(\mathrm{z}_j+s)} 
    = \frac{exp(\mathrm{z}_c)exp(s)}{\sum_{j=1}^D exp(\mathrm{z}_j)exp(s)} 
    = \sigma(\mathrm{z}_c)
\]

Thus, Proposition \ref{prop-logitshift} is proved.
    
\end{proof}

\subsection{Proof of Proposition \ref{prop-unbounded}}\label{Appendix-Proof-unbounded}
\begin{proof}
From Eq.\ref{equa-Energy}, if $s \leq 0$, we have,
\[
    -\infty < -E(\mathbf{z}+s) =  \mathrm{log}\sum_{c=1}^C e^{(\mathrm{z}_c+s)} = \mathrm{log}\sum_{c=1}^C e^{\mathrm{z}_c} +s \leq -E(\mathbf{z})
\]
Similarly, if $s > 0$, we have,
\[
    +\infty > -E(\mathbf{z}+s) =  \mathrm{log}\sum_{c=1}^C e^{(\mathrm{z}_c+s)} =  \mathrm{log}\sum_{c=1}^C e^{\mathrm{z}_c} +s > -E(\mathbf{z})
\]

Thus, Proposition \ref{prop-unbounded} is proved.
    
\end{proof}

\subsection{Proof of Proposition \ref{prop-Energy_bound} } \label{Appendix-proof-Energy_bound}
    
Since the $\mathrm{log}$ function is monotonically increasing, we simplify the problem of deriving an upper bound for the negative energy score $-E$ as follows:
\[
\begin{aligned}
& \text{Maximize} \quad f(\mathbf{z}) = \sum_{c=1}^C e^{\mathrm{z}_c} \\
& \text{Subject to} \quad g(\mathbf{z}) = \sum_{c=1}^C (\mathrm{z}_c)^2 - (M_{\mathrm{norm}})^2 = 0
\end{aligned}
\]

Define the Lagrangian function:
\[
L(\mathbf{z}, \lambda) = f(\mathbf{z}) - \lambda g(\mathbf{z})
\]

We aim to prove that at a stationary point \((\mathbf{z}_0, \lambda_0)\), satisfying the conditions:
\[
\begin{aligned}
& \nabla L(\mathbf{z}_0, \lambda_0) = 0, \\
& g(\mathbf{z}_0) = 0,
\end{aligned}
\]
The Lagrangian function attains an extremum.

\begin{proof}

Taking partial derivatives of the Lagrangian function:
\[
\begin{aligned}
\frac{\partial L}{\partial \mathrm{z}_1} &= \frac{\partial f}{\partial \mathrm{z}_1} - \lambda \frac{\partial g}{\partial \mathrm{z}_1} = 0, \\
\frac{\partial L}{\partial \mathrm{z}_2} &= \frac{\partial f}{\partial \mathrm{z}_2} - \lambda \frac{\partial g}{\partial \mathrm{z}_2} = 0, \\
\vdots\\
\frac{\partial L}{\partial \mathrm{z}_C} &= \frac{\partial f}{\partial \mathrm{z}_C} - \lambda \frac{\partial g}{\partial \mathrm{z}_C} = 0, \\
\frac{\partial L}{\partial \lambda} &= -g(\mathbf{z}_0) = 0.
\end{aligned}
\]

Solving this system of equations gives \((\mathbf{z}_0, \lambda_0)\).

When $\lambda_0=0$,  for any $\mathrm{z}_i \in \mathbf{z}$, we have,
\[
    e^{\mathrm{z}_i} = 0
\]
Which does not hold.

When $\lambda_0 \neq 0$, for any $\mathrm{z}_i \in \mathbf{z}$, we have,
\[
        \lambda = \frac{e^{\mathrm{z}_i}}{2\mathrm{z}_i}
\]
then, for any $\mathrm{z}_j \in \mathbf{z}_0$, we have,
\[
    e^{\mathrm{z}_j} - \frac{\mathrm{z}_j}{\mathrm{z}_i} \cdot e^{\mathrm{z}_i} = 0
\]
When $logit$ $\mathbf{z}_0$ is a positive definite matrix, and
\[
\mathrm{z}_1 = \mathrm{z}_2 = \cdots = \mathrm{z}_C = \sqrt{\frac{M_{\mathrm{norm}}^2}{C}}
\]
We have the following upper bound on the negative energy score:
\[
(-E)^{upper} = \mathrm{log}C + \sqrt{\frac{M_{\mathrm{norm}}^2}{C}}
\]

Similarly, when $logit$ $\mathbf{z}_0$ is a negative definite matrix, and
\[
\mathrm{z}_1 = \mathrm{z}_2 = \cdots = \mathrm{z}_C =- \sqrt{\frac{M_{\mathrm{norm}}^2}{C}}
\]
We have the following lower bound on the negative energy score:
\[
(-E)^{lower} = \mathrm{log}C - \sqrt{\frac{M_{\mathrm{norm}}^2}{C}}
\]

\end{proof}

\subsection{More Analysis on $\mathcal{L}_{\mathrm{UB}}$}\label{Appendix-sec-more_analys_on_Lub}
\subsubsection{How does $\mathcal{L}_{\mathrm{bound}}$ work?}\label{Appendix-subsec-more_analys_on_Lb}
We give the gradient of $\mathcal{L}_{\mathrm{bound}}$ over $\mathrm{logit}$ $\mathbf{z}$ as follows:
\[
\begin{aligned}
\frac{\partial \mathcal{L}_{\mathrm{bound}}}{\partial z_i} 
&= \overbrace{ \frac{1}{M_{\mathrm{norm}}|\mathcal{V}|} \cdot \Big( 2(\lVert \mathbf{z} \rVert_2 - \frac{1}{|\mathcal{V}|} \cdot \sum_{j=1}^{|\mathcal{V}|}\lVert \mathbf{z_j} \rVert_2) \cdot (1-\frac{1}{|\mathcal{V}|}) + \sum_{\mathbf{z}_m \neq \mathbf{z}} 2(\lVert \mathbf{z_m} \rVert_2 - \frac{1}{|\mathcal{V}|} \cdot \sum_{j=1}^{|\mathcal{V}|}\lVert \mathbf{z_j} \rVert_2) \cdot (\frac{-1}{|\mathcal{V}|}) \Big) }^{\frac{\partial \mathcal{L}_{\mathrm{bound}}}{\partial \lVert \mathbf{z} \rVert_2}} \\
&\cdot  \underbrace{\Big( (\sum_{j=1}^C \mathrm{z}_j^2 )^{\frac{-1}{2}} \cdot \mathrm{z}_i \Big) }_{\frac{\partial \lVert \mathbf{z} \rVert_2}{\partial \mathrm{z}_i}}\\
\end{aligned}
\]
We find that when $\lVert \mathbf{z} \rVert_2$ deviates from the mean $\frac{1}{|\mathcal{V}|} \cdot \sum_{j=1}^{|\mathcal{V}|}\lVert \mathbf{z_j} \rVert_2$, $\frac{\partial \mathcal{L}_{\mathrm{bound}}}{\partial \lVert \mathbf{z} \rVert_2}$ produces a gradient that brings $\lVert \mathbf{z} \rVert_2$ closer to the mean. Interestingly, $\frac{\partial \lVert \mathbf{z} \rVert_2}{\partial \mathrm{z}_i}$ can be viewed as a scaling factor that acts with strength proportional to $|\mathrm{z}_i|$.

\subsubsection{How does $\mathcal{L}_{\mathrm{uniform}}$ work?}\label{Appendix-subsec-more_analys_on_Lu}
We give the gradient of $\mathcal{L}_{\mathrm{uniform}}$ over $\mathrm{logit}$ $\mathbf{z}$ as follows:
\[
\begin{aligned}
\frac{\partial \mathcal{L}_{\mathrm{uniform}}}{\partial z_i} 
&= \overbrace{ \frac{1}{M_{\mathrm{sum}}|\mathcal{V}|} \cdot \Big( 2(\Sigma(\mathbf{z}) - \frac{1}{|\mathcal{V}|} \cdot \sum_{j=1}^{|\mathcal{V}|}\Sigma(\mathbf{z_j})) \cdot (1-\frac{1}{|\mathcal{V}|}) + \sum_{\mathbf{z}_m \neq \mathbf{z}} 2(\Sigma(\mathbf{z}_m) - \frac{1}{|\mathcal{V}|} \cdot \sum_{j=1}^{|\mathcal{V}|}\Sigma(\mathbf{z}_j)) \cdot (\frac{-1}{|\mathcal{V}|}) \Big) }^{\frac{\partial \mathcal{L}_{\mathrm{uniform}}}{\partial \Sigma(\mathbf{z})}} \cdot  \overbrace{1}^{\frac{\partial \Sigma(\mathbf{z})}{\partial \mathrm{z}_i}}\\
\end{aligned}
\]
We find that when $\Sigma(\mathbf{z})$ deviates from the mean $\frac{1}{|\mathcal{V}|} \cdot \sum_{j=1}^{|\mathcal{V}|}\Sigma(\mathbf{z_j})$, $\frac{\partial \mathcal{L}_{\mathrm{uniform}}}{\partial \Sigma(\mathbf{z})}$ produces a gradient that brings $\Sigma(\mathbf{z})$ closer to the mean. $\frac{\partial \Sigma(\mathbf{z})}{\partial \mathrm{z}_i}$ shows that the strength of $\mathcal{L}_{\mathrm{uniform}}$'s action is independent of $\mathrm{z}_i$, and shifts $\mathbf{z}$ as a whole. According to Proposition \ref{prop-logitshift}, this shift does not change the supervised classification results.

\section{Experimental Details} \label{Appendix-Experimental Details}
We supplement experiment details for reproducibility. Our implementation is based on Ubuntu 20.04, Cuda 12.1, Pytorch 1.12.0, and Pytorch Geometric 2.1.0.post1. All the experiments run with an NVIDIA 3090 with 24GB memory.

\subsection{Dataset Information} \label{Appendix-Dataset Information}
The datasets used in our experiment are publicly available as standard benchmarks for evaluating graph learning models. For ogbn-Arxiv, we utilize the preprocessed dataset and data loader provided by the OGB package. We use the data loader provided by the Pytorch Geometric package for other datasets.

\begin{itemize}

\item \textit{Cora} dataset, widely used in graph-based machine learning, is a citation network often applied to node classification and link prediction tasks. It provides a rich representation of scholarly connections with 2,708 nodes, 5,429 edges, 1,433 features, and seven classes. Following the GNNSAFE \citep{wu2023energy}, we employ three approaches to synthetically generate Out-of-Distribution (OOD) data, as outlined in Section \ref{subsec-Setup}, where under the Label Leave-out, we take four classes of the samples as ID data and three classes as OOD data. For training/validation/testing, we adopt a semi-supervised learning approach inspired by \citep{kipf2016semi}, utilizing the specified data splits on the in-distribution data.

% Our split for the Cora, Twitch-Explicit, and ogbn-arxiv datasets is consistent with GNNSAFE \citep{wu2023energy}. The split for the Citeseer and Pubmed datasets is as follows:

\item \textit{Citeseer} dataset is a citation network commonly applied to node classification and link prediction tasks. It comprehensively represents citation relationships with 3,327 nodes, 4,732 edges, 3,703 features, and six classes. We use three strategies to generate synthetic OOD data (see Section \ref{subsec-Setup} for details), where under the Label Leave-out setting, we take four classes of the samples as ID data and two as OOD data. For training/validation/testing, we adhere to the semi-supervised learning setting introduced by \citep{kipf2016semi}, utilizing the provided data splits on the in-distribution data.

\item \textit{PubMed} dataset is another citation network commonly employed in graph-based machine learning and network analysis tasks, focusing on biomedical literature. The Pubmed dataset comprises 19,717 nodes, 44,338 edges, 500 features, and three classes. The nodes represent scientific papers, and the features encompass bag-of-words representations of the paper content. We use three strategies to generate synthetic OOD data (see Section \ref{subsec-Setup} for details), where under the Label Leave-out, we take two classes of the samples as ID data and one class as OOD data. For training/validation/testing, we adhere to the semi-supervised learning setting introduced by \citep{kipf2016semi}, utilizing the provided data splits on the in-distribution data.

\item \textit{Twitch-Explicit} dataset comprises multiple graphs, with each subgraph representing a distinct social network in a specific region. Within this dataset, nodes correspond to game players on Twitch, and edges signify mutual follow-up relationships between users. Node features include embeddings representing the games played by Twitch users, while the label class indicates whether a user streams mature content. Subgraph sizes vary, ranging from 1,912 to 9,498 nodes, with edge numbers spanning 31,299 to 153,138. A shared feature dimension of 2,545 is maintained across all subgraphs. Following the methodology outlined in GNNSAFE \citep{wu2023energy} and detailed in Section \ref{subsec-Setup}, we designate subgraph DE as in-distribution data, employing random splits with a ratio of 1:1:8 for training, validation, and testing. Additionally, subgraph EN serve as out-of-distribution (OOD) exposure, while ES, FR, and RU are utilized for OOD testing data.

\item \textit{ogbn-Arxiv} dataset is a comprehensive graph dataset that captures citation information from 1960 to 2020. Each node represents a paper, with its subject area serving as the label for prediction. Edges in the graph denote citation relationships, and every node is associated with a 128-dimensional feature vector derived from the word embeddings of its title and abstract. As outlined in Section \ref{subsec-Setup}, we utilize time information to partition the in-distribution and out-of-distribution  data. For the in-distribution subset, we employ random splits in a 1:1:8 ratio for training, validation, and testing. Our partitioning approach for the in-distribution and OOD data differs from the original splitting strategy proposed by \citep{hu2020open}, which relied on-time information to split training and test nodes. In our configuration, the random splitting ensures that the training, validation, and testing nodes (of in-distribution data) are drawn from an identical distribution distinct from the OOD data.

\end{itemize}

\subsection{Implementation Details} \label{Appendix-Implementation Details}
\paragraph{Architecture}
% Our basic encoder backbone is a GCN-based encoder, and we also consider other models such as MLP, GAT, JKNet, and MixHop for further discussion. We summarize the architectural information of all the encoder backbones used in our experiments as
% follows:
% \begin{itemize}
%     \item GCN: 2 GCNConv layers with hidden size 64, ReLU activation, self-loop, and batch normalization.
%     \item MLP: 2 fully connected layers with hidden size 64 and ReLU activation.
%     \item GAT: 2 GATConv layers with hidden size 64, ELU activation, head number [2, 1], and batch normalization.
%     \item JKNet: 2 GCNConv layers with hidden size 64, ReLU activation, self-loop, and batch normalization. The jumping knowledge module uses max pooling.
%     \item MixHop: 2 MixHop layers with hop number 2, hidden size 64, ReLU activation, and batch normalization.
% \end{itemize}
Our primary encoder backbone is a Graph Convolutional Network (GCN)-based encoder. Additionally, we explore other models, including Multi-Layer Perceptron (MLP), Graph Attention Network (GAT), Jumping Knowledge Network (JKNet), and MixHop, for further investigation. We summarize the architectural details of all the encoder backbones used in our experiments as follows:

\begin{itemize}
    \item \textbf{GCN}: This architecture comprises 2 GCNConv layers with a hidden size 64, ReLU activation function, self-loop mechanism, and batch normalization.
    
    \item \textbf{MLP}: The MLP architecture consists of 2 fully connected layers with a hidden size 64 and ReLU activation function.
    
    \item \textbf{GAT}: The GAT model includes 2 GATConv layers with a hidden size 64, ELU activation function, head number [2, 1], and batch normalization.
    
    \item \textbf{JKNet}: This architecture includes 2 GCNConv layers with a hidden size 64, ReLU activation function, self-loop mechanism, and batch normalization. The jumping knowledge module utilizes max pooling.
    
    \item \textbf{MixHop}: The MixHop model comprises 2 MixHop layers with a hop number of 2, a hidden size 64, a ReLU activation function, and batch normalization.
\end{itemize}
\paragraph{Hyper-parameters} 
We Follow the GNNSAFE\citep{wu2023energy} to set the hyper-parameters $m_{\mathrm{in}}$, $m_{\mathrm{out}}$, $\alpha$, $\eta$ and $K$. Additionally, we consider grid-search for $\lambda_1 \in \{0.0001, 0.001,0.01,0.1,1\}$ and $\lambda_2 \in \{0.01, 0.1, 1, 10,100\}$, and we use $\lambda_1 = 0.001$ and $\lambda_2 = 1$ as the default setting.

\paragraph{Training Details}
In each run, we train the model with 200 epochs as a fixed budget and report the testing performance produced by the
epoch where the model yields the lowest classification loss on validation data.

\paragraph{Competitors}
We use the results reported in the \citep{wu2023energy} for the baseline MSP, ODIN, Mahalanobis, Energy and Energy FT, GKDE, GPN, GNNSAFE, and GNNSAFE++. We use the hyperparameters recommended in it for LogitNorm \citep{liu2020energy}.

\subsection{Evaluation Metrics} \label{Appendix-Evaluation Metrics}
We provide more details concerning the evaluation metrics we used in our experiment. The metrics FPR95, AUROC (Area Under the Receiver Operating Characteristic curve), and AUPR (Area Under the Precision-Recall curve) are commonly used in binary classification tasks to assess the performance of machine learning models. 
\begin{itemize}
\item FPR95 is a specific operating point on the ROC curve. It represents the false positive rate when the true positive rate is fixed at 95\%. In other words, it indicates the rate of false positives when the model achieves a high level of sensitivity (95\% true positive rate). 
\item AUROC measures the area under the ROC curve, a graphical representation of the trade-off between true positive rate (sensitivity) and false positive rate (1-specificity) for different classification thresholds. An AUROC score of 1.0 indicates a perfect classifier, while a score of 0.5 corresponds to a random classifier. 
\item AUPR measures the area under the precision-recall curve, which plots precision (positive predictive value) against recall (sensitivity) for different classification thresholds. AUPR is particularly useful when the data is imbalanced, as it focuses on the classifier's performance in the positive class. Similar to AUROC, a higher AUPR indicates better classifier performance.
\end{itemize}

\section{Additional Experiment Results} \label{Appendix-Additional Experiment Results}
We supplement more experimental results in this section. In specificin Table \ref{table-A3}, \ref{table-A4} and \ref{table-A5}, we report the AUROC/AUPR/FPR95 and in-distribution testing accuracy on Cora, Citeseer, and Pubmed as complementary for Table \ref{tabel-Cora_C_P} in the main text. Besides, we report detailed OOD detection performance on each OOD dataset (different subgraphs for Twitch and other years for Arxiv) in Table \ref{table-A1} and \ref{table-A2} complementary for Table \ref{tabel-Twitch_A} in the main text. 

Table \ref{table-train_time} compares the models' training time (TR) per epoch and inference time (IN) of all the models. Our implementation is based on Ubuntu 20.04, Cuda 12.1, Pytorch 1.12.0, and Pytorch Geometric 2.1.0.post1. All the experiments run with an NVIDIA 3090 with 24GB memory.  The results indicate that the additional training cost brought by our methods compared to the previous SOTA method GNNSAFE is almost negligible. 

Additionally, we conduct experiments comparing the convergence speed of the models. Table \ref{table-convergence_speed} demonstrates that compared to GNNSAFE, our NODESAFE exhibits significantly faster convergence speed and better OOD detection performance. This indicates that $\mathcal{L}_{\mathrm{bound}}$ and $\mathcal{L}_{\mathrm{uniform}}$ regularization can allow the energy of ID and OOD to reach a stable distribution more quickly, while also significantly enhancing the ability of GNN to detect OOD nodes.

\begin{figure}[h]
  \centering
  \includegraphics[width=1\textwidth]{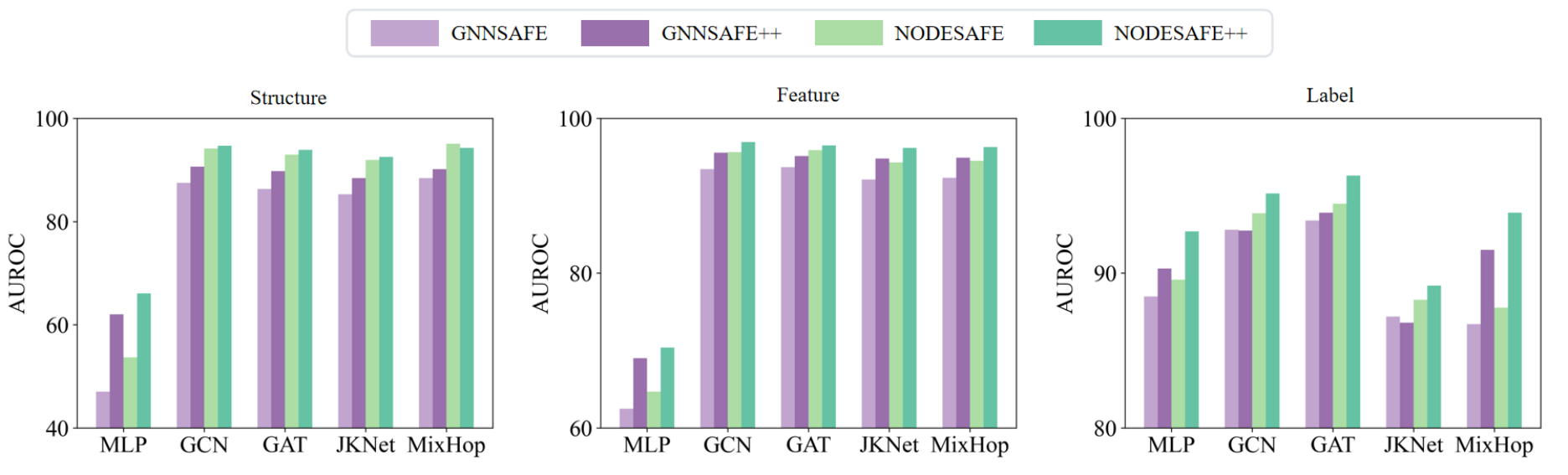}
  \caption{Performance comparison of GNNSAFE, GNNSAFE++, NODESAFE and NODESAFE++ w.r.t. different encoder backbones on \textit{Cora} with three OOD types.}
  \label{F-backbone}
\end{figure}

Figure \ref{F-backbone} compares the performance of GNNSAFE, GNNSAFE++, NODESAFE, and NODESAFE++ w.r.t. the use of different encoder backbones including MLP, GCN, GAT, JKNet, and MixHop for the three OOD types of Cora. We observe that the relative performance of the five models is generally consistent with the results in Table 1.
% We use GCN as the backbone, and the two NODESAFE models outperform the competitors in almost all cases.

\begin{table*}[h]
% \scriptsize
\centering
\caption{OOD detection results measured by AUROC ($\uparrow$) / AUPR ($\uparrow$) / FPR95 ($\downarrow$) on \textit{Twitch}.}\label{table-A1}
\resizebox{\linewidth}{!}{
\begin{tabular}{c|c|ccc|ccc|ccc}
\hline
\hline
\multirow{2}*{Model}  &OOD &\multicolumn{3}{c|}{Twitch-ES} &\multicolumn{3}{c|}{Twitch-FR} &\multicolumn{3}{c}{Twitch-RU}\\
~ &Expo &AUROC(↑)	&AUPR(↑)	&FPR95(↓)&AUROC(↑)	&AUPR(↑)	&FPR95(↓)		&AUROC(↑)	&AUPR(↑)	&FPR95(↓)	\\
\hline
\hline
MSP	&No	&37.72	&53.08	&98.09	&	21.82&	38.27&	99.25	&	41.23	&56.06&	95.01\\
ODIN	&No &83.83	&80.43	&33.28		&59.82	&64.63	&92.57		&58.67	&72.58&	93.98\\
Mahalanobis&No	&45.66	&58.82	&95.48	&	40.40&	46.69	&95.54	&	55.68	&66.42&	90.13\\
Energy	&No	&38.80	&54.23	&95.70	&	57.21&	61.48	&91.57&		57.72&	66.68	&87.57\\
GKDE	&No	&48.70	&61.05	&95.37	&	49.19	&52.94&	95.04	&	46.48&	62.11	&95.62\\
GPN	&No	&53.00	&64.24	&95.05	&	51.25	&55.37	&93.92	&	50.89&	65.14	&99.93\\
GNNSAFE	&No	&49.07	&57.62&	93.98		&63.49&	66.25	&90.80		&87.90	&89.05	&43.95\\
GNNSAFE w/ $\mathcal{L}_{LN}$ 	&No &60.87	&74.94	&93.72		&52.15	&58.63	&96.52	&	59.49&	71.25	&92.11\\
NODESAFE	&No	&94.44	&95.78	&25.84		&96.12&	97.30	&24.93	&	79.40	&86.91	&90.24\\
\hline							
OE	&Yes	&55.97	&69.49	&94.94		&45.66	&54.03	&95.48		&55.72	&70.18	&95.07\\
Energy FT	&Yes	&80.73	&87.56&	76.76		&79.66	&81.20	&76.39		&93.12	&95.36	&30.72\\
GNNSAFE++	&Yes	&94.54	&97.17&	44.06	&	93.45&	95.44	&51.06		&98.10	&98.74	&5.59\\
GNNSAFE++   w/ $\mathcal{L}_{LN}$ 	&Yes&	96.70	&98.43	&24.35		&91.20	&94.64	&68.63	&	98.09&	99.09	&8.44\\
NODESAFE++	&Yes	&97.66	&98.86	&3.10		&97.94&	98.75&	7.86		&99.75	&99.88	&0.05\\
\hline
\hline
\end{tabular}
}
\end{table*}

\begin{table*}[h]
% \scriptsize
\centering
\caption{OOD detection results measured by AUROC ($\uparrow$) / AUPR ($\uparrow$) / FPR95 ($\downarrow$) on \textit{ogbn-Arxiv}.}\label{table-A2}
\resizebox{\linewidth}{!}{
\begin{tabular}{c|c|ccc|ccc|ccc}
\hline
\hline
\multirow{2}*{Model}  &OOD &\multicolumn{3}{c|}{Arxiv-2018} &\multicolumn{3}{c|}{Arxiv-2019} &\multicolumn{3}{c}{Arxiv-2020}\\
~ &Expo &AUROC(↑)	&AUPR(↑)	&FPR95(↓)&AUROC(↑)	&AUPR(↑)	&FPR95(↓)		&AUROC(↑)	&AUPR(↑)	&FPR95(↓)	\\
\hline
\hline
MSP	&No&	61.66&	70.63	&91.67		&63.07	&66.00	&90.82	&	67.00	&90.92&	89.28 \\
ODIN	&No	&53.49	&63.06	&100.00	&	53.95	&56.07	&100.00		&55.78	&87.41	&100.00 \\
Mahalanobis	&No	&57.08	&65.09	&93.69	&	56.76&	57.85&	94.01	&	56.92	&85.95	&95.01 \\
Energy	&No	&61.75	&70.41	&91.74		&63.16	&65.78	&90.96	&	67.70	&91.15	&89.69 \\
GKDE	&No	&56.29	&66.78	&94.31		&57.87	&62.34	&93.97	&	60.79	&88.74	&93.31 \\
GPN	&No	&----	&----	&&----		&----	&----&	----		&----	----	&---- \\
GNNSAFE&	No&	66.47	&74.99	&89.44	&	68.36&	71.57&	88.02	&	78.35&	94.76&	83.57 \\
GNNSAFE w/ $\mathcal{L}_{LN}$	&No&	66.89&	75.31	&88.33		&68.91&	71.97&	86.95		&78.72&	94.84	&82.52 \\
NODESAFE	&No&	67.90	&76.33	&87.30		&69.85	&73.15	&85.38		&79.59	&95.08	&80.10 \\
\hline												
OE	&Yes	&67.72&	75.74	&86.67		&69.33	&72.15&	85.52		&72.35&	92.57	&83.28 \\
Energy FT	&Yes	&69.58	&76.31	&82.10		&70.58	&72.03	&81.30		&74.53	&93.08&	78.36 \\
GNNSAFE++	&Yes	&70.40&	78.62	&81.47	&	72.16	&75.43&	79.33	&	81.75	&95.57&	71.50 \\
GNNSAFE++ w/ $\mathcal{L}_{LN}$&	Yes&	67.73&	76.40	&87.86		&69.58	&73.27	&86.81		&79.33	&95.04	&81.79 \\
NODESAFE++	&Yes	&71.13&	79.19&	79.73	&	72.92	&76.18	&77.30		&82.42	&95.73	&68.71 \\
\hline
\hline
\end{tabular}
}
\end{table*}

\begin{table*}[h]
% \scriptsize
\centering
\caption{OOD detection results measured by AUROC ($\uparrow$) / AUPR ($\uparrow$) / FPR95 ($\downarrow$) on \textit{Cora}.}\label{table-A3}
\resizebox{\linewidth}{!}{
\begin{tabular}{c|c|cccc|cccc|cccc}
\hline
\hline
\multirow{2}*{Model}  &OOD &\multicolumn{4}{c|}{Cora-Structure} &\multicolumn{4}{c|}{Cora-Feature} &\multicolumn{4}{c}{Cora-Label}\\
~ &Expo &AUROC(↑)	&AUPR(↑)	&FPR95(↓) &ID ACC(↑) &AUROC(↑)	&AUPR(↑)	&FPR95(↓)	&ID ACC(↑)	&AUROC(↑)	&AUPR(↑)	&FPR95(↓)	&ID ACC(↑)\\
\hline
\hline
MSP&	No	&70.90&	45.73&	87.30	&75.50	&	85.39&	73.70	&64.88	&75.30	&	91.36	&78.03	&34.99	&88.92\\
ODIN	&No&	49.92	&27.01	&100.00	&74.90		&49.88	&26.96	&100.00	&75.00		&49.80	&24.27	&100.00	&88.92\\
Mahalanobis	&No	&46.68	&29.03	&98.19	&74.90		&49.93	&31.95	&99.93&	74.90	&	67.62	&42.31	&90.77	&88.92\\
Energy	&No&	71.73	&46.08&	88.74	&76.00		&86.15	&74.42	&65.81&	76.10	&	91.40	&78.14	&41.08	&88.92\\
GKDE	&No	&68.61	&44.26	&84.34	&73.70		&82.79	&66.52	&68.24	&74.80		&57.23	&27.50	&88.95	&89.87\\
GPN	&No	&77.47&	53.26	&76.22	&76.50		&85.88	&73.79	&56.17	&77.00		&90.34	&77.40 &37.42	&91.46\\
GNNSAFE	&No	&87.52&	77.46&	73.15&	75.80		&93.44	&88.19	&38.92&	76.40		&92.80	&82.21	&30.83&	88.92\\
GNNSAFE w/ $\mathcal{L}_{LN}$	&No	&88.33	&78.13&	61.04	&80.00		&93.26	&86.89	&38.44	&80.80		&93.50	&85.19	&34.99&	89.24\\
NODESAFE	&No	&94.07&	83.98	&25.63	&77.20		&95.30&	88.82&	23.08	&78.70		&93.80	&85.22&	29.41	&89.87\\
\hline												
OE	&Yes	&67.98	&46.93&	95.31	&71.80		&81.83	&70.84	&83.79	&73.30		&89.47	&77.01	&46.55	&87.97\\
Energy FT	&Yes&	75.88	&49.18	&67.73&	75.50	&	88.15	&75.99	&47.53&	75.30	&	91.36	&78.49	&37.83	&90.51\\
GNNSAFE++	&Yes&	90.62	&81.88	&53.51&	76.10		&95.56&	90.27	&27.73&	76.80		&92.75	&82.64	&34.08	&91.46\\
GNNSAFE++ w/ $\mathcal{L}_{LN}$&	Yes&	90.13	&81.61	&51.99&	79.70		&94.11&	88.82	&32.72	&79.70		&93.83	&84.17&	28.40	&89.56\\
NODESAFE++	&Yes&	94.64	&85.63	&23.34&	76.40	&	96.56&	91.96	&14.73	&77.10	&	94.88&	86.66	&22.52	&91.46\\
\hline
\hline
\end{tabular}
}
\end{table*}

\begin{table*}[h]
% \scriptsize
\centering
\caption{OOD detection results measured by AUROC ($\uparrow$) / AUPR ($\uparrow$) / FPR95 ($\downarrow$) on \textit{Citeseer}.}\label{table-A4}
\resizebox{\linewidth}{!}{
\begin{tabular}{c|c|cccc|cccc|cccc}
\hline
\hline
\multirow{2}*{Model}  &OOD &\multicolumn{4}{c|}{Citeseer-Structure} &\multicolumn{4}{c|}{Citeseer-Feature} &\multicolumn{4}{c}{Citeseer-Label}\\
~ &Expo &AUROC(↑)	&AUPR(↑)	&FPR95(↓) &ID ACC(↑) &AUROC(↑)	&AUPR(↑)	&FPR95(↓)	&ID ACC(↑)	&AUROC(↑)	&AUPR(↑)	&FPR95(↓)	&ID ACC(↑)\\
\hline
\hline
MSP&	No&	66.34	&34.78	&85.03	&65.60	&	78.32&	55.48&	71.27&	66.20	&	88.42&	64.03	&51.97&	89.36\\
ODIN	&No	&49.23	&23.07&	100.00&	66.10	&	49.86&	23.11	&100.00&	65.80	&	51.33	&17.97&	100.00	&89.36\\
Mahalanobis	&No	&45.26&	21.20	&99.13	&60.70	&	49.92	&31.20	&99.73	&53.30		&53.46	&35.47&	86.32&	72.51\\
Energy	&No	&65.62	&33.63	&87.59	&65.20		&79.19	&55.94	&69.67	&64.50	&	89.98&	64.10	&38.76	&90.58\\
GKDE	&No	&61.48&	31.55	&93.71	&64.70		&74.68	&50.25	&71.22	&64.20		&82.69&	61.21	&50.61	&89.16\\
GPN&	No	&70.55	&41.12	&78.26	&65.80		&78.46	&53.21	&73.14&	63.20		&85.65	&62.32	&41.37&	89.30\\
GNNSAFE	&No	&79.79	&60.81	&74.72&	64.90	&	83.46&	67.02	&68.83	&64.40	&	90.01	&65.26	&36.53	&90.58\\
GNNSAFE w/ $\mathcal{L}_{LN}$	&No	&84.67	&69.73&	76.35	&65.40		&88.11	&76.20	&72.35	&65.40		&90.47	&67.69	&37.32	&86.93\\
NODESAFE	&No	&88.40	&75.93&	57.89&	69.90		&90.41	&79.30	&45.47	&68.60		&91.66&	68.15	&29.30	&90.58\\
\hline															
OE&	Yes	&58.74&	30.07&	95.37	&59.00		&72.06&	48.80&	81.09&	60.50		&89.44	&62.74&	45.99&	87.23\\
Energy FT	&Yes&	68.87	&36.01	&76.44&	63.00		&79.23	&55.69	&64.08	&64.40		&91.34&	66.66&	31.60	&90.58\\
GNNSAFE++	&Yes	&82.43	&65.58&	70.72	&65.90		&83.27&	68.06	&72.98&	65.10		&91.57	&65.48	&29.30	&88.15\\
GNNSAFE++ w/ $\mathcal{L}_{LN}$	&Yes	&84.93	&70.90	&74.81	&65.50		&87.68	&76.10&	75.47&	65.30	&	91.00&	67.57&	30.55	&84.50\\
NODESAFE++&	Yes	&86.90&	71.41	&52.60&	65.00		&91.14	&79.48	&40.49	&66.50	&	91.98	&68.97	&29.04	&88.15\\
\hline
\hline
\end{tabular}
}
\end{table*}

\begin{table*}[h]
% \scriptsize
\centering
\caption{OOD detection results measured by AUROC ($\uparrow$) / AUPR ($\uparrow$) / FPR95 ($\downarrow$) on \textit{Pubmed}.}\label{table-A5}
\resizebox{\linewidth}{!}{
\begin{tabular}{c|c|cccc|cccc|cccc}
\hline
\hline
\multirow{2}*{Model}  &OOD &\multicolumn{4}{c|}{Pubmed-Structure} &\multicolumn{4}{c|}{Pubmed-Feature} &\multicolumn{4}{c}{Pubmed-Label}\\
~ &Expo &AUROC(↑)	&AUPR(↑)	&FPR95(↓) &ID ACC(↑) &AUROC(↑)	&AUPR(↑)	&FPR95(↓)	&ID ACC(↑)	&AUROC(↑)	&AUPR(↑)	&FPR95(↓)	&ID ACC(↑)\\
\hline
\hline
MSP	&No	&74.31	&17.44	&84.08	&75.10		&83.28&	39.29	&69.38	&75.00		&85.71	&34.98	&46.19	&100.00\\
ODIN	&No	&49.76	&4.83&	100.00	&75.30		&49.67	&4.83&	100.00&	75.30		&56.24	&13.49&	100.00	&100.00\\
Mahalanobis	&No	&55.28	&8.38&	97.59	&69.30		&69.12	&15.09	&84.93	&73.00		&75.77&	23.40	&78.21&	97.52\\
Energy	&No	&74.33&	17.32	&78.90	&75.60		&84.16&	39.10	&62.47&	75.50	&	86.81	&36.00&	45.14	&100.00\\
GKDE	&No	&74.02&	16.89	&81.52&	75.20	&	82.25	&32.41&	68.56	&74.10		&83.36	&34.63	&69.52&	100.00\\
GPN	&No	&74.96&	17.54	&80.33&	75.80		&82.56	&39.75	&61.79	&74.50		&86.51&	35.12	&50.23&	100.00\\
GNNSAFE	&No	&87.52	&62.74&	44.64	&75.80		&94.28&	71.66	&33.89	&76.40	&	88.02	&44.77	&36.49	&100.00\\
GNNSAFE w/ $\mathcal{L}_{LN}$	&No	&89.31	&58.72&	75.10	&74.90	&	92.07	&64.21&	49.93	&73.00		&91.12	&54.33	&32.29	&93.61\\
NODESAFE	&No&	94.13	&71.29	&23.80	&77.00	&	95.97	&78.22	&22.01&	77.60	&	93.80	&71.98&	25.01	&100.00\\
\hline															
OE&	Yes	&74.41	&16.74&	83.52	&72.90	&	82.34&	38.60	&74.58&	73.10	&	81.97	&29.88&	60.30	&99.02\\
Energy FT&	Yes&	73.54	&18.00	&92.04	&75.80		&78.95	&37.21	&90.00	&75.30		&91.83	&52.39	&25.59&	100.00\\
GNNSAFE++	&Yes	&90.62&	72.78	&34.43	&76.10		&95.16&	77.47&	26.30	&76.80	&	87.98	&41.43	&33.63	&96.07\\
GNNSAFE++ w/ $\mathcal{L}_{LN}$	&Yes&	86.21	&57.25	&91.58&	73.30	&	87.56&	58.53	&86.17&	72.90	&	89.66&	44.87	&27.81	&91.89\\
NODESAFE++	&Yes&	96.30&	81.88	&14.52	&74.50	&	95.26&	78.12	&24.45	&77.30	&	93.48	&53.45&	23.81&	100.00\\
\hline
\hline
\end{tabular}
}
\end{table*}

\begin{table*}[h]
% \scriptsize
\centering
\caption{Comparison of training time per epoch and inference time of all the models. Our implementation is based on Ubuntu 20.04, Cuda 12.1, Pytorch 1.12.0, and Pytorch Geometric 2.1.0.post1. All the experiments run with an NVIDIA 3090 with 24GB memory.}\label{table-train_time}
\resizebox{\linewidth}{!}{
\begin{tabular}{c|cc|cc|cc|cc|cc}
\hline
\hline
\multirow{2}*{Model}  &\multicolumn{2}{c|}{Cora} &\multicolumn{2}{c|}{Citeseer} &\multicolumn{2}{c}{Pubmed} &\multicolumn{2}{c}{Twitch} &\multicolumn{2}{c}{Arxiv}\\
~ &TR(s) &IN(s)	&TR(s) &IN(s) &TR(s) &IN(s) &TR(s) &IN(s) &TR(s) &IN(s)\\
\hline
\hline
MSP	&0.006	&0.003	&0.007	&0.003	&0.019	&0.053	&0.006	&0.029	&0.040	&0.161\\
ODIN	&0.006	&0.005	&0.007	&0.006	&0.017	&0.099	&0.006	&0.059	&0.017	&0.009\\
Mahalanobis	&0.007	&0.047	&0.008	&0.053	&0.023	&0.343	&0.007	&0.004	&0.022	&0.032\\
Energy	&0.008	&0.010	&0.009	&0.011	&0.036	&0.051	&0.008	&0.035	&0.059	&0.179\\
GKDE	&0.007	&0.015	&0.008	&0.017	&0.022	&0.075	&0.004	&0.031	&0.028	&0.159\\
GPN	&7.336	&25.730	&8.290	&29.075	&9.071	&36.322	&2.081	&10.132	&-	&-\\
OE	&0.011	&0.012	&0.012	&0.014	&0.037	&0.052	&0.009	&0.032	&0.072	&0.234\\
Energy FT	&0.012	&0.003	&0.014	&0.003	&0.042	&0.054	&0.012	&0.038	&0.064	&0.203\\
GNNSAFE	&0.008	&0.004	&0.009	&0.005	&0.037	&0.052	&0.008	&0.036	&0.058	&0.204\\
GNNSAFE++	&0.012	&0.004	&0.014	&0.005	&0.045	&0.060	&0.012	&0.038	&0.070	&0.222\\
NODESAFE (ours)	&0.008	&0.004	&0.009	&0.005	&0.039	&0.054	&0.009	&0.037	&0.060	&0.206\\
NODESAFE++ (ours)	&0.013	&0.004	&0.015	&0.005	&0.049	&0.062	&0.014	&0.040	&0.073	&0.233\\
\hline
\hline
\end{tabular}
}
\end{table*}

\begin{table*}[h]
% \scriptsize
\centering
\caption{Comparison of Convergence Speed. The epoch with the minimum loss on the validation set is considered the best epoch (the smaller the epoch, the faster the convergence speed), and its corresponding AUROC (the higher the better) is recorded.}\label{table-convergence_speed}
\resizebox{\linewidth}{!}{
\begin{tabular}{c|cc|cc|cc|cc|cc}
\hline
\hline
\multirow{2}*{Model}  &\multicolumn{2}{c|}{Cora} &\multicolumn{2}{c|}{Citeseer} &\multicolumn{2}{c}{Pubmed} &\multicolumn{2}{c}{Twitch} &\multicolumn{2}{c}{Arxiv}\\
~ &Best Epoch(↓) &AUROC(↑) 	&Best Epoch(↓) &AUROC(↑) &Best Epoch(↓) &AUROC(↑) &Best Epoch(↓) &AUROC(↑) &Best Epoch(↓) &AUROC(↑)\\
\hline
\hline
GNNSAFE	&148	&87.52	&123	&79.79	&167	&87.52	&156	&66.82	&145	&71.06\\
NODESAFE	&61	&94.07	&66	&88.40	&76	&94.13	&87	&89.99	&111	&72.44\\
\hline
GNNSAFE++	&176	&81.88	&157	&65.58	&165	&72.78	&169	&95.36	&154	&74.77\\
NODESAFE++	&56	&94.64	&97	&86.90	&119	&96.30	&98	&98.50	&127	&75.49\\
\hline
\hline
\end{tabular}
}
\end{table*}

\end{document}